\documentclass[runningheads]{llncs}
\usepackage{graphicx}
\usepackage[misc]{ifsym}
\usepackage{bbding}
\usepackage{xspace}

\usepackage{tikz}
\usepackage{amsmath,amssymb} 
\usepackage{color}
\usepackage{booktabs}
\usepackage{pifont}
\usepackage{multirow}
\usepackage[font={small}]{caption}
\usepackage{wrapfig} 

\usepackage{orcidlink}

\makeatletter
\DeclareRobustCommand\onedot{\futurelet\@let@token\@onedot}
\def\@onedot{\ifx\@let@token.\else.\null\fi\xspace}

\def\eg{\emph{e.g}\onedot} 
\def\ie{\emph{i.e}\onedot}

\makeatother

\begin{document}
\pagestyle{headings}
\mainmatter
\def\ECCVSubNumber{709}  

\title{CelebV-HQ: A Large-Scale Video Facial Attributes Dataset}

\titlerunning{CelebV-HQ}
%
\def\thefootnote{*}\footnotetext{Equal contribution.}\def\thefootnote{\arabic{footnote}}
\def\thefootnote{\Letter}\footnotetext{Corresponding author (wuwenyan0503@gmail.com).}\def\thefootnote{\arabic{footnote}}
\author{
Hao Zhu\inst{1*}\orcidlink{0000-0003-2155-1488} \and
Wayne Wu\inst{1*}\textsuperscript{\Letter}\orcidlink{0000-0002-1364-8151} \and 
Wentao Zhu\inst{2}\orcidlink{0000-0002-5483-0259} \and
Liming Jiang\inst{3}\orcidlink{0000-0001-8109-5598} \and \\
Siwei Tang\inst{1}\orcidlink{0000-0002-3105-551X} \and
Li Zhang\inst{1}\orcidlink{0000-0002-8714-2137} \and 
Ziwei Liu\inst{3}\orcidlink{0000-0002-4220-5958} \and
Chen Change Loy\inst{3}\orcidlink{0000-0001-5345-1591}}
\authorrunning{H. Zhu et al.}

\institute{SenseTime Research \and
Peking University \and
S-Lab, Nanyang Technological University
}

\maketitle 
\begin{center}
  \centering
  \captionsetup{type=figure}
  \includegraphics[width=1\textwidth]{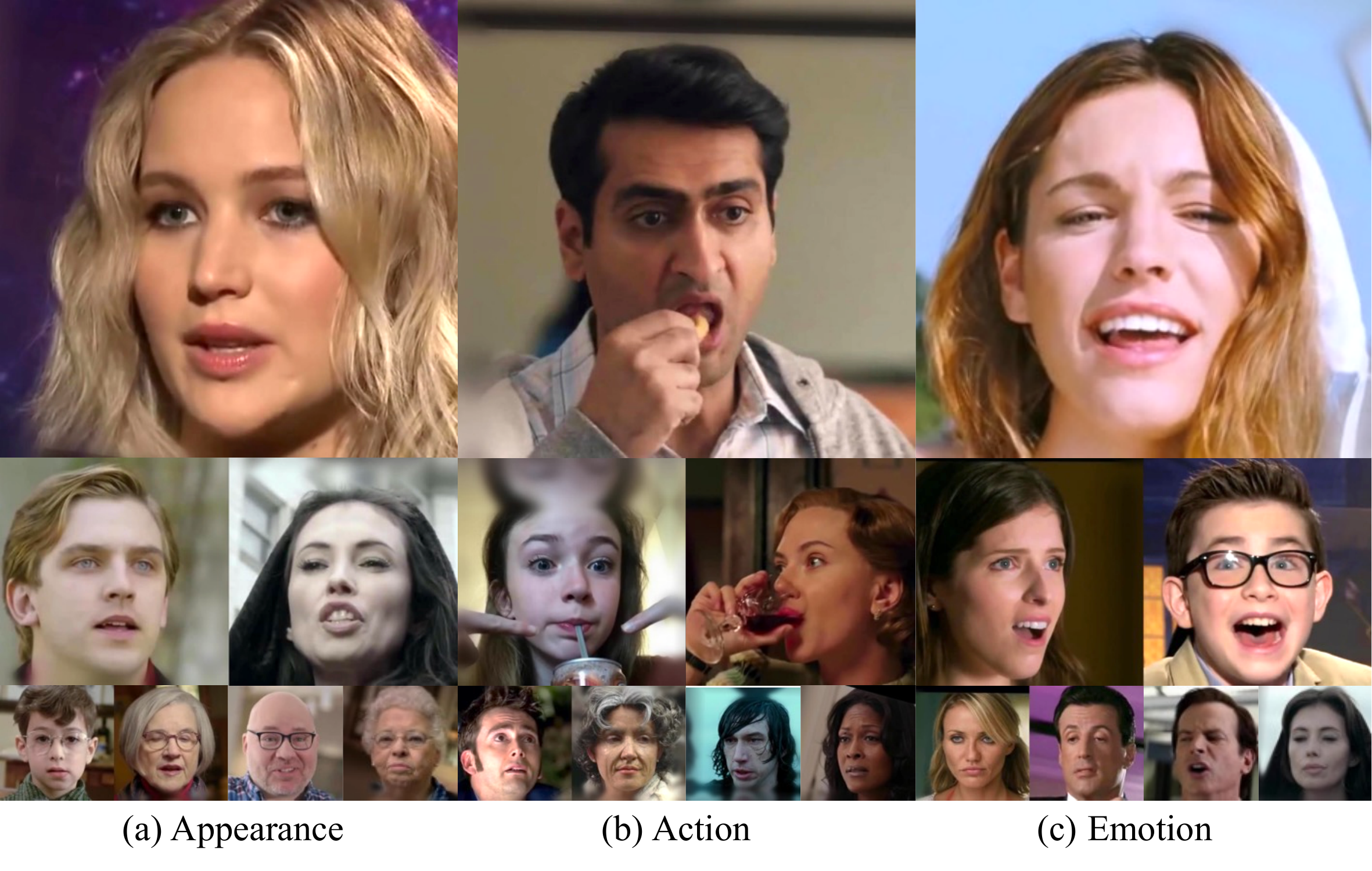}
  \caption{\textbf{Overview of CelebV-HQ.} CelebV-HQ contains $35,666$ videos, including $15,653$ identities. Each video was manually labeled with $83$ facial attributes, covering appearance, action, and emotion attributes.} 
  \label{fig:teaser}
\end{center}%

\begin{abstract}
\vspace{-3mm}
Large-scale datasets have played indispensable roles in the recent success of face generation/editing and significantly facilitated the advances of emerging research fields.
However, the academic community still lacks a video dataset with diverse facial attribute annotations, which is crucial for the research on face-related videos.
In this work, we propose a large-scale, high-quality, and diverse video dataset with rich facial attribute annotations, named the High-Quality Celebrity Video Dataset (CelebV-HQ). CelebV-HQ contains $35,666$ video clips with the resolution of $512\times512$ at least, involving $15,653$ identities. All clips are labeled manually with $83$ facial attributes, covering appearance, action, and emotion.
We conduct a comprehensive analysis in terms of age, ethnicity, brightness stability, motion smoothness, head pose diversity, and data quality to demonstrate the diversity and temporal coherence of CelebV-HQ. Besides, its versatility and potential are validated on two representative tasks, \ie, unconditional video generation and video facial attribute editing.
Furthermore, we envision the future potential of CelebV-HQ, as well as the new opportunities and challenges it would bring to related research directions. Data, code, and models are publicly available\footnote{Project page: \url{https://celebv-hq.github.io/} \\
Code and models: \url{https://github.com/CelebV-HQ/CelebV-HQ}}.

\end{abstract}


\section{Introduction}
\begin{sloppypar}
    
The rapid development of Generative Adversarial Networks (GANs)~\cite{gan,dcgan,celebahq,stylegan,stylegan2,stylegan3} has demonstrably promoted advances in face generation and editing.
This progress relies heavily on the contribution of large-scale datasets, \eg, CelebA~\cite{celeba15}, CelebA-HQ~\cite{celebahq} and FFHQ~\cite{stylegan}.
These datasets, with high-quality facial images, have facilitated the development of a series of face generation and editing tasks, such as unconditional face generation~\cite{gan,dcgan,stylegan,stylegan2,mocoganhd,digan,styleganv,DBLP:conf/nips/JiangDWL21}, facial attribute editing~\cite{starganv2,munit,interfacegan,e4e,xu2022transeditor} and neural rendering~\cite{headnerf,stylenerf,or2021stylesdf,chan2021pigan,or2021stylesdf,eg3d,guo2021adnerf,Gafni_2021_CVPR,chen2022sem2nerf}.
However, most of these efforts are based on static \textit{image modality}. In industry, with the booming development of mobile internet~\cite{iotsurvey} and mobile phone~\cite{yang2019analysis}, \textit{video modality} data begins to take a bigger and bigger share in customers' daily shootings~\cite{snapchat,tiktok}. A well-suited dataset, which is capable of supporting the face generation and editing tasks in video modality, is eagerly asked.
\end{sloppypar}


Recent works~\cite{stylegan2,brock2018large} have shown that the \textit{scale} and \textit{quality} are essential factors for a facial dataset in image modality. A more sufficient utilization of large-scale datasets would improve model generalization~\cite{schmidt2018adversarially}, while the quality of the dataset largely determines the limit of the generative models~\cite{dcgan,stylegan,stylegan2,mocoganhd,digan,styleganv}. In addition, facial \textit{attribute} provides effective information to help researchers go more deeply into the face-related topics~\cite{celeba15,munit,starganv2,interfacegan}. However, the current public facial datasets consist of either static images with attribute labels~\cite{celeba15,celebahq} or videos with insufficient scale~\cite{mead} and quality~\cite{vox17,vox2}.


Constructing a large-scale and high-quality face video dataset with diverse facial attribute's annotations is still an open question, given the challenges brought by the nature of video data.
1) Scale. The collected videos need to meet several requirements, such as temporal consistency, high-resolution and full-head. The strict standards together with the limited sources, make the expansion of dataset's scale both time and labor consuming.
2) Quality. The quality is not only reflected in the high fidelity and resolution, but also in the diverse and natural distribution of data samples. It asks for a well-designed data-filtering process to ensure all of the requirements of fidelity, resolution and data distribution.
3) Attribute Annotation. The coverage of the facial attribute set need to be sufficient to  describe a human face thoroughly, both in the time-invariant and time-variant perspective. Also, the annotation process need to be accurate and highly efficient.


In order to tackle the challenges discussed above, we carefully device a procedure for dataset construction.
%
First, to ensure the scale of the collected video, we build a large and diverse set of Internet queries that are built from the Wikipedia dataset~\cite{wikidataset}, covering $11$ languages, with 8376 entities, and 3717 actions. The designed queries cover a rich set of scenarios and thus successfully enable a huge raw data pool with millions of clips.
Then, to filter out high-quality data from the raw data pool, we introduce an automatic pre-processing pipeline. In this pipeline, we leverage face detection and alignment tools to ensure the high fidelity and resolution. Also, with the huge raw data pool, after the pre-processing, we are still able to get a dataset with real-world distributions with large diversity.
Finally, we propose a facial attributes set with extensive coverage, including appearance, action and emotion.
To ensure the accuracy and efficiency of the annotation, we design a systematic attributes annotation process, including annotator training, automatic judgment and quality check steps.

To this end, we successfully conduct the High-Quality Celebrity Video (CelebV-HQ) Dataset, a large-scale, diverse, and high-quality video facial dataset with abundant attributes' annotations.
CelebV-HQ contains $35,666$ in-the-wild video clips with the resolution of $512\times512$ at least, involving $15,653$ person identities and $83$ manually labeled facial attributes.
Our careful labeling comprises a comprehensive set of face-related attributes, including $40$ appearance attributes, $35$ action attributes, and $8$ emotion attributes. Samples on CelebV-HQ are shown in Fig.~\ref{fig:teaser}.

After constructing the CelebV-HQ, we perform a comprehensive analysis of data distribution to demonstrate its statistical superiority to both image and video datasets. 
First, compared to image datasets with attribute annotations~\cite{celeba15,celebahq}, CelebV-HQ has much higher resolution ($2\times$) than CelebA~\cite{celeba15} and comparable scale to high-quality dataset~\cite{celebahq}. Also, by comparing CelebV-HQ with CelebA-HQ~\cite{celebahq} in the \textit{time-invariant} aspects, we demonstrate that CelebV-HQ has a reasonable distribution on appearance and facial geometry.
Furthermore, we compare CelebV-HQ with a representative video face dataset VoxCeleb2~\cite{vox2} in the \textit{time-variant} aspects, such as temporal data quality, brightness variation, head pose distribution, and motion smoothness, which verify that CelebV-HQ has superior video quality.

Besides, to demonstrate the effectiveness and potential of CelebV-HQ, we evaluate representative baselines in two typical tasks: unconditional video generation and video facial attribute editing.
For the task of unconditional video generation, we train state-of-the-art unconditional video GANs~\cite{mocoganhd,digan} on CelebV-HQ fullset and its subsets that divided by different actions. When trained on different subsets of CelebV-HQ, the corresponding actions can be successfully generated.
Further, we explore the video facial attribute editing task using temporal constrained image-to-image baselines~\cite{munit,starganv2}. Thanks to the rich sequential information included in CelebV-HQ dataset, we show that a simple modification of current image-based methods can bring remarkable improvement in the temporal consistency of generated videos. The experiments conducted above empirically demonstrate the effectiveness of our proposed CelebV-HQ dataset.

In addition to face video generation/editing tasks we evaluated in this work, CelebV-HQ could potentially benefit the academic community in many other fields. For example, neural rendering (\eg, novel view synthesis~\cite{headnerf,stylenerf,or2021stylesdf} and 3d face generation~\cite{chan2021pigan,or2021stylesdf,eg3d,guo2021adnerf,Gafni_2021_CVPR,chen2022sem2nerf}) and face analysis (\eg, attribute recognition~\cite{zhong2016face,ding2018deep,fairface}, action recognition~~\cite{wang2013action,jegham2020vision}, emotion recognition~~\cite{dzedzickis2020human,lee2019context} and forgery detection~\cite{li2020face,haliassos2021lips,zhu2021face}). We finally provide several empirical insights during constructing CelebV-HQ dataset and make an exhaustive discussion of the potential of CelebV-HQ in research community.

In summary, our contributions are threefold:
1) We contribute the first large-scale face video dataset, named CelebV-HQ, with high-quality video data and diverse manually annotated attributes. Corresponding to CelebA-HQ~\cite{celebahq}, CelebV-HQ fills in the blank on video modality and facilitates future research.
2) We perform a comprehensive statistical analysis in terms of attributes diversity and temporal statistics to show the superiority of CelebV-HQ.
3) We conduct extensive experiments on typical video generation/editing tasks, which demonstrates the effectiveness and potential of CelebV-HQ.

\section{Related Work}

\subsection{Video Face Generation and Editing}
\begin{sloppypar}

Recent advances in face video generation typically focused on unconditional video generation~\cite{vgan,tgan,mocogan,digan,mocoganhd,styleganv} and conditional face video generation~\cite{wayne2018reenactgan,fomm,bilayermodel,wang2021facevid2vid,zhou2019talking,chen2019hierarchical,zhu2021arbitrary,zhou2021pose,ji2021audio}. 
Conventional unconditional video face generation~\cite{vgan,tgan,mocogan,digan} are based on GANs~\cite{gan}. 
These models usually decompose the latent code into content and motion codes to control the corresponding signals. 
Some recent efforts~\cite{mocoganhd,styleganv} aimed to extend high-quality pre-trained image generators to a video version to exploit the rich prior information. 
Conditional face video generation mainly including face reenactment~\cite{wayne2018reenactgan,fomm,bilayermodel,wang2021facevid2vid} and talking face generation~\cite{zhou2019talking,chen2019hierarchical,zhu2021arbitrary,zhou2021pose,ji2021audio}. The motivation of these tasks is to use visual and audio modalities to guide the motion of a face video.
\end{sloppypar}

Face video editing is another emerging field~\cite{yao2021latent,tzaban2022stitch}. 
The common characteristic of these works is to edit face attributes on the StyleGAN~\cite{stylegan2} latent space. 
Nevertheless, due to the lack of large-scale high-quality video datasets, these video-based editing efforts are still trained on images, by exploiting the rich information of a pre-trained image model ~\cite{stylegan2}. This leads to the main problem they have to solve to improve temporal consistency.
Therefore, with such a face video dataset proposed, there is much room for improvement in current video face generation and editing methods.

\subsection{Face Datasets} 
Face datasets can be divided into two categories: image datasets and video datasets. 
First, many face image datasets are initially proposed for face recognition, like LFW~\cite{LFW} and CelebFaces~\cite{celebfaces} which largely promote the development of related fields.
To analyze facial attributes, datasets like CelebA and LFWA \cite{celeba15} have been proposed. Both of them have 40 facial attribute annotations and have advanced the research field to a finer level of granularity. CelebA-HQ~\cite{celebahq} improves 30k images in CelebA to 1024$\times$1024 resolution. CelebAMask-HQ~\cite{celebahq_mask} further labels 19 classes of segmentation masks. CelebA-Dialog~\cite{celeba_dialog} labels captions describing the attributes.  
In addition to the above image datasets, many video datasets have also been released. CelebV~\cite{wayne2018reenactgan} was proposed for face reenactment, and it contains five long videos collected from the Internet. 
Audiovisual datasets such as VoxCeleb~\cite{vox17} and VoxCeleb2~\cite{vox2} were originally released for speaker recognition, and further stimulated the development of audiovisual speaker separation and talking face generation domains. 
MEAD~\cite{mead} is the largest emotional video dataset, which includes 60 actors recorded from seven view directions.
However, all of these datasets either contains only images with attribute annotations or are unlabeled videos with insufficient diversity. The rapidly growing demand for video facial attribute editing cannot be met. A video version of the dataset like CelebA-HQ \cite{celebahq} is urgently needed to be proposed.

\section{CelebV-HQ Construction}
The dataset lies the foundation for model training, and its quality greatly affects the downstream tasks. The principle of building CelebV-HQ is to reflect real-world distribution with large-scale, high-quality, and diverse video clips. Hence, we design a rigorous and efficient pipeline to construct CelebV-HQ dataset, including Data Collection, Data Pre-processing, and Data Annotation.

\begin{figure*}[t]
\centering
\includegraphics[width=1\textwidth]{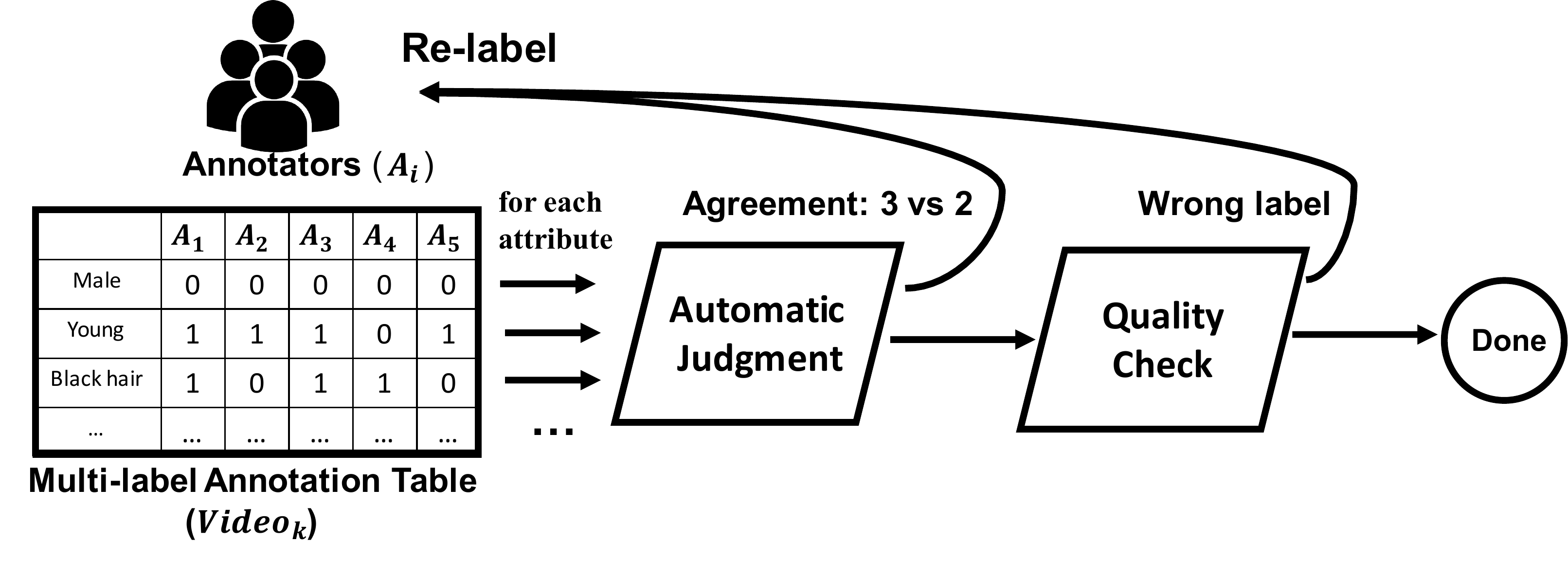}\caption{\textbf{The pipeline of data annotation.} Each video clip is annotated by $5$ annotators. When there is a large discrepancy in the annotations or the labeling results do not meet the standard of professional quality inspectors, these samples will be re-labeled.}
\label{fig:data_pipeline}
\vspace{-2mm}
\end{figure*}
\subsection{Data Collection}

The data collection process consists of the following steps. 
We start by creating various queries in order to retrieve human videos that are diverse in content and rich in attributes. The queries are designed to include keywords of different categories such as celebrity names, movie trailers, street interviews and vlogs, all in different languages. 
Then, we use these queries to collect the raw videos from the Internet. During the collection process, we have several constraints to filter the unsatisfactory videos. For each query, we only collect the first $30$ results to reduce duplicate human IDs, and the raw videos are required to have a resolution greater than $1080$p with a normal bitrate. Consequently, we obtain a raw data pool with millions of video clips.

\subsection{Data Pre-processing}
\label{sec:video_preprocess}
In order to sample high-quality face video clips from the raw data pool, we develop an automatic video pre-processing pipeline. The pipeline is illustrated in Appendix~\ref{appsec:video_preprocess}.
First, we detect $98$ facial landmarks~\cite{wu2018look} from the video, and use these to extract the bounding boxes. Faces smaller than $450$ pixels are filtered out. 
Then, we check whether the adjacent frames belong to the same person based on the motion~\cite{sort} and identity~\cite{arcface}. If not, we will split the video into different clips.
Next, given a sequence of bounding boxes, we calculate their minimum bounding rectangle. To reduce data loss, we expand the bounding rectangle smaller than $512^2$ to this size, and use the bounding box to crop the original video.
Finally, only clips longer than 3 seconds are kept.

We choose $512^2$ as the normalized resolution due to the following reasons. 
1) The face regions of web videos usually do not reach the resolution of $1024^2$ or higher, and it is difficult to obtain super high-resolution videos and ensure their diversity. Before the rescaling, the percentage of video resolution: $0.6\%$ for $450^2{\sim}512^2$, $76.6\%$ for $512^2{\sim}1024^2$, and $22.7\%$ for $1024^2+$.
2) We need to make sure that all the videos are at the same resolution when training models. We choose $512^2$ to ensure that all the clips are not upsampled significantly, which would affect the video quality.

Also, to meet different usage scenario, a tool is provided on our project page that offers options to keep the original resolution.

\subsection{Data Annotation}
Data annotation is a core part of CelebV-HQ, and the annotation accuracy is vital. We first describe how we select the attributes to be annotated, then present the standard protocol of manual annotation.

\noindent
\textbf{Attribute Selection.} We decouple a face video into three factors, \ie,  appearance, action, and emotion. Appearance describes the facial attributes that do not change along with the video sequence, such as hair color and gender. Action describes facial attributes that are related with video sequence, such as laugh and talk. Emotion describes the high-level mental status of human, such as neutral and happy.
These three categories serve as important feature dimensions to characterize face video clips.

For appearance attributes, we derive most of the classes from CelebA~\cite{celeba15}. However, we find that three common attributes (\ie, ``long hair'', ``sunglasses'' and ``wearing a mask'') in real-world videos are not defined in CelebA~\cite{celeba15}. We add these three attributes to the appearance attributes as well.
Meanwhile, some action-related attributes, such as ``smiling'' and ``mouth slightly open'', have been removed. This process yields $40$ appearance attributes in total.
For action attributes, inspired by Kinetics-700~\cite{kinetics700}, we select the face-related actions from its classes and add other facial actions from Internet tags to ensure that the final 35 attributes could cover common facial actions. For emotion attributes, we follow the $8$ emotions designed in RAVESS~\cite{RAVDESS}, including neutral, anger, contempt, disgust, fear, happiness, sadness, and surprise. Note that the appearance and action attributes are all multi-label as the classes are not mutually exclusive, while emotion attributes are designed to be single-label. We append the complete list of all the attributes to the Appendix~\ref{sec:attribute_list}.

\noindent
\textbf{Attribute Annotation.}
To ensure the accuracy of the annotations, our entire annotation process includes the training of annotators, annotation, and quality control. 
Before the labeling begins, training courses are provided to help annotators understand each attribute and to have the same criteria for judging each attribute. 
As shown in Fig.~\ref{fig:data_pipeline}, we first set up a Multi-label Annotation Table for each video, the table contains all labels that need to be labeled. Each video clip is independently annotated by $5$ trained annotators. 
We design an Automatic Judgment strategy, when there is a discrepancy in the annotations of an attribute, we select the annotation that has been agreed the most. If the annotation is only marginally agreed ($3$ vs $2$), the sample will be re-labeled.
Finally, we additionally take a Quality Check process, in which the annotated data is further inspected by a professional quality inspector with research experiments in Computer Vision. If the annotated data does not meet the standard, it will also be re-labeled.

\section{Statistics}

In this section, we present the statistics of CelebV-HQ to demonstrate its statistical superiority. 
Then, we make comparison of CelebV-HQ with two most related and representative image and video datasets (\ie, CelebA-HQ~\cite{celebahq} and VoxCeleb2~\cite{vox2}) respectively, in which we verify that the proposed CelebV-HQ has a natural distribution and better quality.

\begin{table*}[!h]
\footnotesize
\caption{\textbf{Face datasets comparison.} The symbol ``\#'' indicates the number. The abbreviations ``Id.'', ``Reso.'', ``Dura.'', ``App.'', ``Act.'', ``Emo.'', ``Env.'', and ``Fmt.'' stand for Identity, Resolution, Duration, Appearance, Action, Emotion, Environment, and Format, respectively. The ``$*$'' denotes the estimated resolution. }
\centering
\resizebox{1\linewidth}{!}
{
\begin{tabular}{l|cccc|ccc|c|c}
\hline
          & \multicolumn{4}{c|}{\textbf{Meta Infomation}}        & \multicolumn{3}{c|}{\textbf{Attributes}}                        & \multirow{2}{*}{\textbf{Env.}} & \multirow{2}{*}{\textbf{Fmt.}} \\ \cline{2-8}
 \textbf{Datasets} & \#Samples & \#Id. & Reso. & Dura. & App. & Act. & Emo. & & \\ \hline
CelebA~\cite{celeba15}    &   202,599 & 10,177 &   178$\times$218    & N/A & \textcolor{red}{\ding{51}} & \ding{55} & \ding{55} & Wild & IMG \\
CelebA-HQ~\cite{celebahq} & 30,000 &      6,217 &  1024$\times$1024  &   N/A       &     \textcolor{red}{\ding{51}}    & \ding{55}  &\ding{55}  & Wild & IMG    \\
FFHQ~\cite{stylegan} & 70,000 &      N/A &  1024$\times$1024  &   N/A       &     \ding{55}    & \ding{55}  &\ding{55}  & Wild & IMG    \\
\hline
CelebV~\cite{wayne2018reenactgan}  &  5      &     5         &    256$\times$256    &   2hrs       & \ding{55}       & \ding{55}                    & \ding{55}                     & Wild        & VID                   \\ 
FaceForensics~\cite{faceforensics}  &  1,004      &     1,004         &    256$\times$256$*$    &   4hrs       & \ding{55}       & \ding{55}                    & \ding{55}                     & Wild        & VID                   \\ 
VoxCeleb~\cite{vox17} &     21,245      &       1,251       &  224$\times$224     &    352hrs      & \ding{55}       & \ding{55}                    & \ding{55}                     & Wild           & VID                   \\
VoxCeleb2~\cite{vox2} &        150,480    &  6,112  &    224$\times$224   &   2,442hrs       &  \ding{55}       & \ding{55} & \ding{55} & Wild & VID    \\
MEAD~\cite{mead}  &  281,400      &     60         &    1980$\times$1080    &   39hrs       & \ding{55}       & \ding{55}                    & \textcolor{red}{\ding{51}}                     & Lab        & VID                   \\ \hline
\textbf{CelebV-HQ}      &   35,666        &   15,653           &  512$\times$512     &    68hrs      & \textcolor{red}{\ding{51}} & \textcolor{red}{\ding{51}} & \textcolor{red}{\ding{51}} & Wild           & VID                   \\ \hline
\end{tabular}
}
\label{tbl:stat}
\end{table*}

\subsection{Analysis of CelebV-HQ}
CelebV-HQ consists of $35,666$ video clips of $3$ to $20$ seconds each, involving $15,653$ identities, with a total video duration of about $65$ hours. For the facial attributes, CelebV-HQ not only has a wide distribution of time-invariant attributes (\ie, appearance), but also has rich samples of time-variant attributes (\ie, action and emotion).

As shown in Table~\ref{tbl:stat}, compared to the image datasets that contain facial attribute annotations~\cite{celeba15,celebahq}, the resolution of CelebV-HQ is more than twice that of CelebA~\cite{celebahq}, and has a comparable scale to the high-quality dataset, CelebA-HQ~\cite{celeba15}. 
More importantly, CelebV-HQ, as a video dataset, contains not only appearance attribute annotations, but also action and emotion attribute annotations, which make it contains richer information than image datasets. %
Other than the diverse annotations, compared to the recent in-the-wild video datasets (CelebV~\cite{wayne2018reenactgan}, FaceForensics~\cite{faceforensics}, VoxCeleb~\cite{vox17} and VoxCeleb2~\cite{vox2}), CelebV-HQ has a much higher resolution.
Specifically, VoxCeleb2~\cite{vox2} and MEAD~\cite{mead}, as two representative face video datasets, are the largest audiovisual video face datasets under in-the-wild and lab-controlled environments respectively.
Althgough the data volume of VoxCeleb2~\cite{vox2} and MEAD~\cite{mead} is relatively large, the videos on these two datasets are homogeneous and in limited distributions. The videos on VoxCeleb2 are mainly talking face, while MEAD was collected in a constrained laboratory environment. In contrast, CelebV-HQ is collected in real-world scenarios with a diverse corpus, making it more natural and rich in the distribution of attributes.

\begin{figure*}[t]
\centering
\includegraphics[width=1\textwidth]{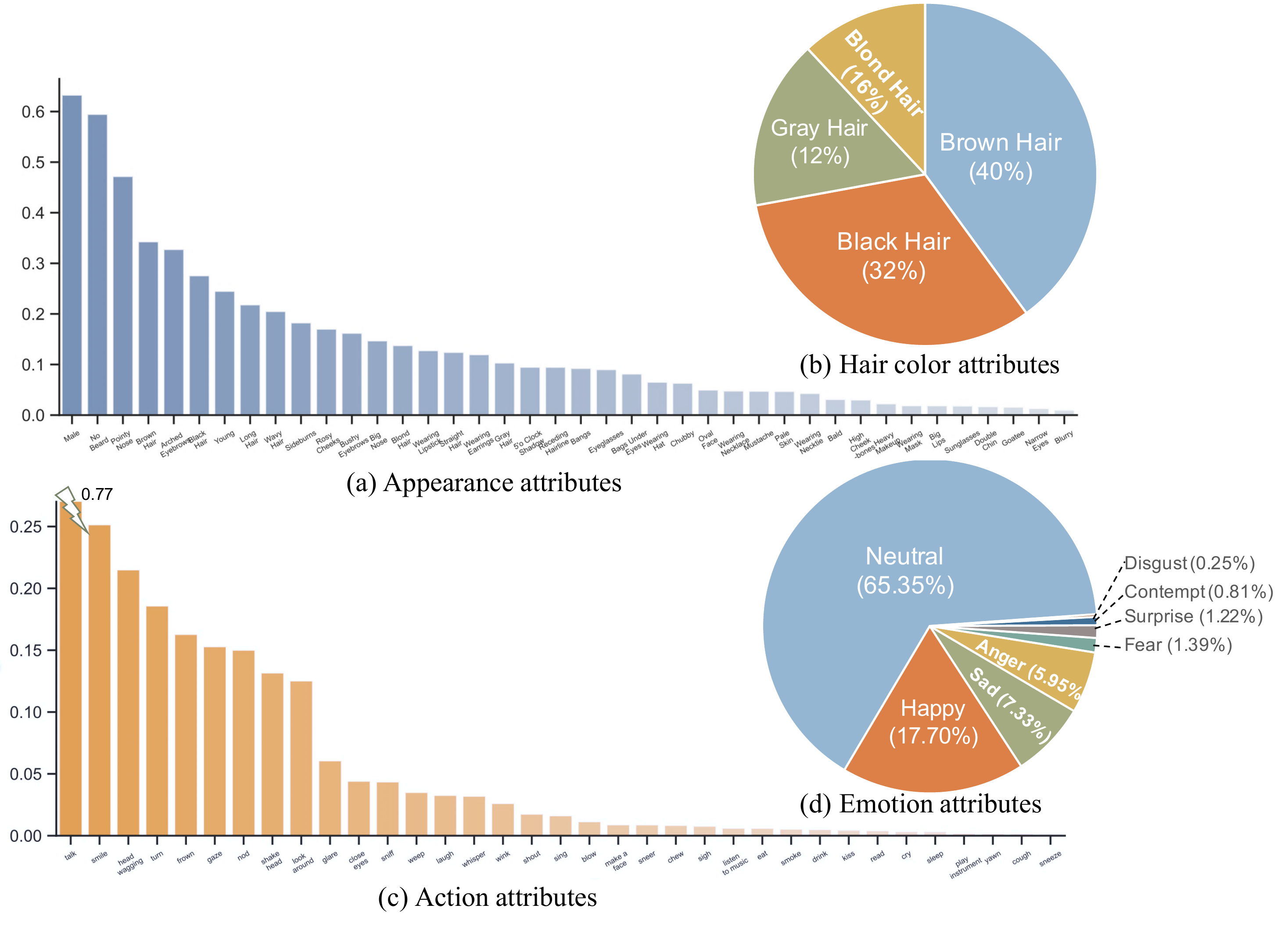}
\caption{\textbf{The distributions of each attribute.} CelebV-HQ has a diverse distribution on each attribute category. (Please zoom in for details).}
\label{fig:pie_attr}
\end{figure*}

We start our analysis of CelebV-HQ with the attribute distribution. 
1) CelebV-HQ contains a total of 40 appearance attributes, as shown in Fig.~\ref{fig:pie_attr}~(a), of which 10 attributes account for more than 20\% each, while there are more than 10 attributes accounting for about 10\% each.
Meanwhile, the overall attribute distribution has a long tail, with 10 attributes accounting for less than 3\% each. 
We compare the hair colors separately, as they are mutually exclusive. From Fig.~\ref{fig:pie_attr}~(b), the distribution in hair color is even, and there are no significant deviations. 
2) There are diverse action attributes in CelebV-HQ as shown in Fig.~\ref{fig:pie_attr}~(c). The common actions, such as ``talk'', ``smile'', and ``head wagging'', account for over 20\% each. About 20 uncommon actions, such as ``yawn'', ``cough'' and ``sneeze'', account for less than 1\% each. This result is in line with our expectation that these uncommon attributes remain open challenges for the academic community. 
3) The proportion of emotion attributes also varies as shown in Fig.~\ref{fig:pie_attr}~(d), with ``neutral'' accounting for the largest proportion, followed by ``happiness'' and ``sadness'' emotions. Unlike the data collected in the laboratory, we do not strictly control the proportion of each attribute, so the overall distribution is more in line with the natural distribution. 
Overall, the CelebV-HQ is a real-world dataset with diverse facial attributes in a \textit{natural distribution}, bringing new opportunities and challenges to the community.

\subsection{Comparison with Image Dataset}
\begin{figure*}[t]
\centering
\includegraphics[width=1\textwidth]{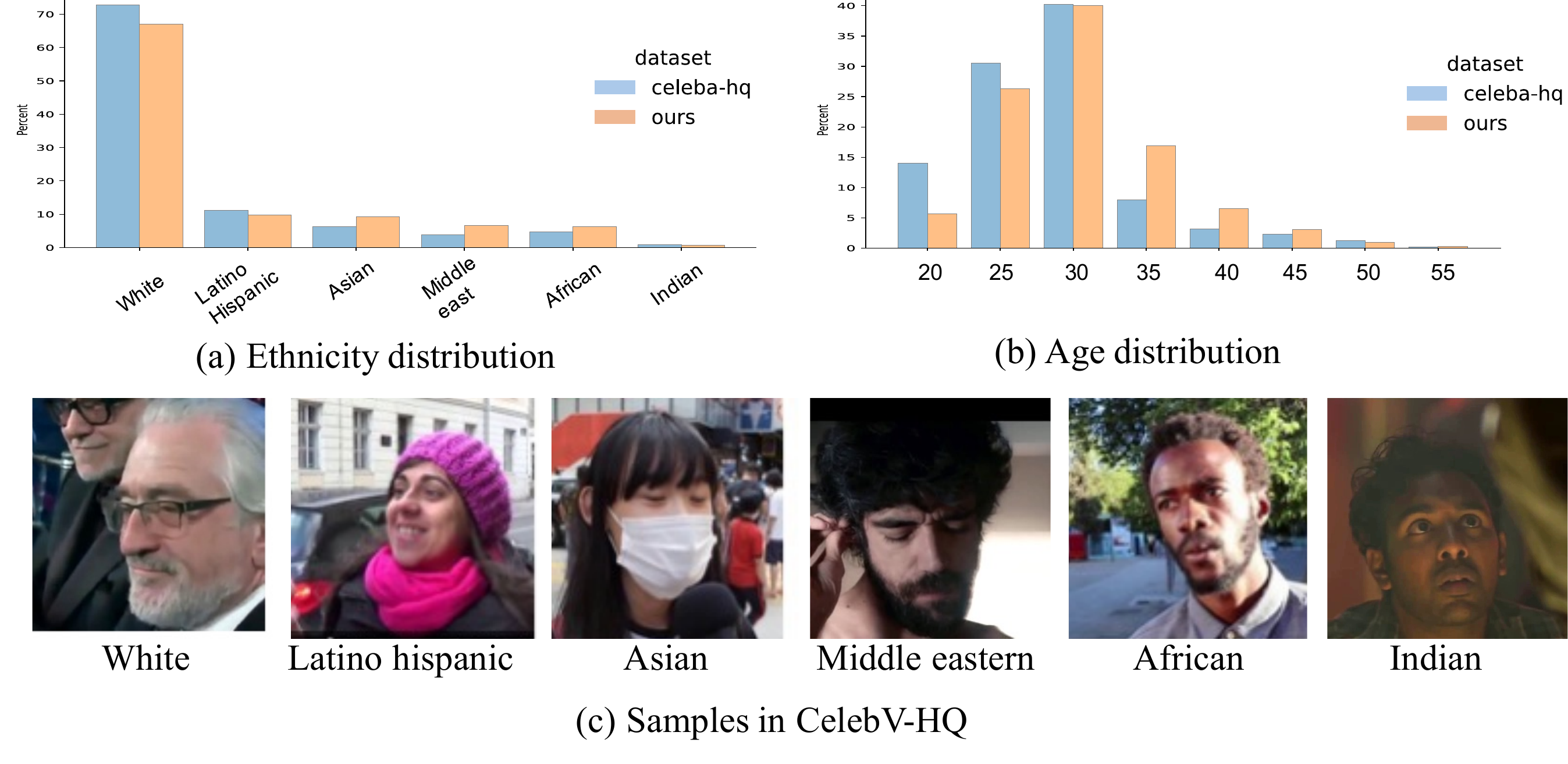}
\caption{\textbf{Distributions of age and ethnicity compared with CelebA-HQ~\cite{celebahq}.} (a) and (b) show that CelebV-HQ has a similar distribution compared to CelebA-HQ~\cite{celebahq}.   
}
\label{fig:attr_dist}
\end{figure*}

Due to CelebV-HQ can be considered as a video version of CelebA-HQ~\cite{celebahq} which is a commonly used dataset and its facial attributes annotation is successful in many works~\cite{shen2017learning,e4e,Yao_2021_ICCV,han2017heterogeneous,ding2018deep,stylegan,stylegan2,starganv2}. We argue that a face video dataset that has similar distribution with CelebA-HQ would also effective. 

To show a reasonable distribution of CelebV-HQ, we compare the proposed CelebV-HQ with CelebA-HQ~\cite{celebahq} in face attribute aspects such as age, ethnicity, and face shape. These factors reflect the basic face information in terms of facial appearance and geometry.
Since ethnicity and age attributes are not explicitly labeled, we estimate them for both datasets using an off-the-shelf facial attribute analysis framework~\cite{serengil2021lightface}.

\noindent
\textbf{Age distribution.} We evaluate whether the dataset is biased towards certain age groups. From Fig.~\ref{fig:attr_dist}~(a), we can see that the age in CelebA-HQ is mainly distributed below 35 years old, while the age distribution of CelebV-HQ is smoother.

\noindent
\textbf{Ethnic Distribution.} The ethnic distribution roughly reflects the data distribution in terms of geography. 
As shown in Fig.~\ref{fig:attr_dist}~(b), CelebV-HQ achieves a distribution close to CelebA-HQ~\cite{celebahq}, and has a more even distribution in Latino Hispanic, Asian, Middle-eastern, and African. 
As shown in Fig.~\ref{fig:attr_dist}~(c), we show a random sample of each ethnic group in the CelebV-HQ.

\begin{wrapfigure}{r}{0.5\textwidth}
\vspace{-5mm}
\centering
\includegraphics[width=0.45\textwidth]{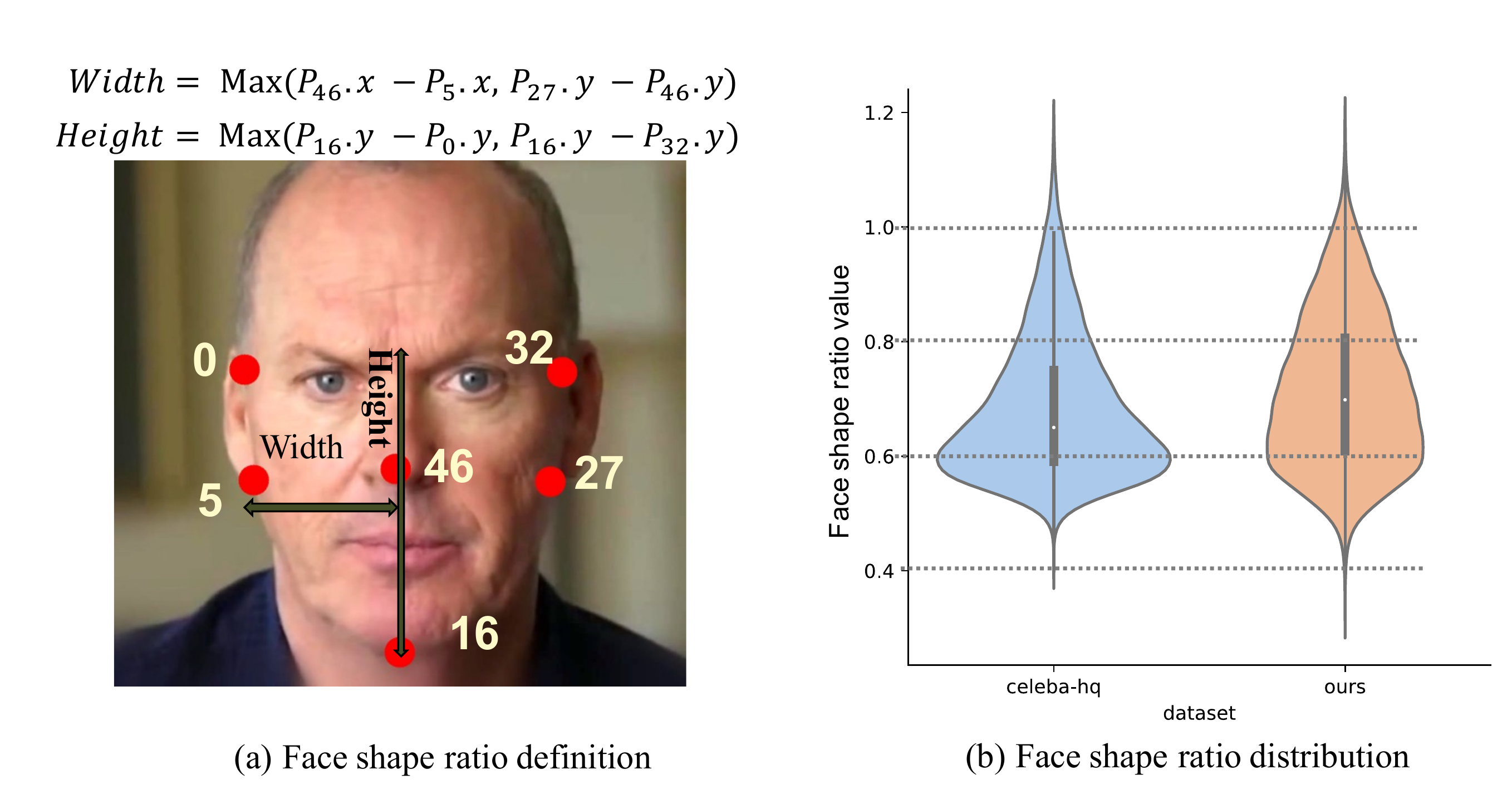}
\caption{\textbf{Distributions of head pose and face shape ratio compared with CelebA-HQ~\cite{celebahq}.} CelebV-HQ contains more diverse head pose and face shape ratio distribution. } 
\label{fig:celeba_headpose_ratio}
\vspace{-5mm}
\end{wrapfigure} 

\noindent \textbf{Face Shape Ratio.} 
In addition, the distribution of face shape ratio indicates the diversity of the dataset in terms of face types. Therefore, a simple analysis of face shape is proposed, where we calculate the ratio of a face using key points~\cite{wu2018look} as shown in Fig.~\ref{fig:celeba_headpose_ratio}~(a).
The distance from the left and right of the cheeks to the nose is recorded as the width, and the distance from the highest point of the cheeks to the chin is recorded as the height. The width-to-height ratio is used as the definition of the face shape ratio.
As shown in Fig.~\ref{fig:celeba_headpose_ratio}~(b), CelebV-HQ has a more uniform distribution, which indicates that the samples in it have diverse face types.

\begin{figure*}[t]
\centering
\includegraphics[width=1\textwidth]{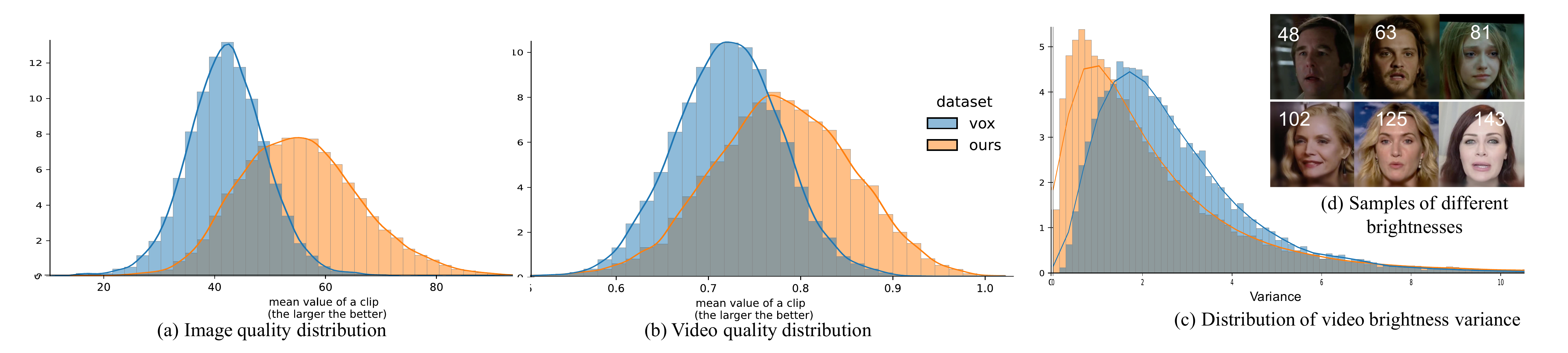}
\caption{\textbf{Distributions of image and video quality, and brightness variance.} (a) Image quality and (b) video quality are measured by BRISQUE~\cite{brisque} and VSFA~\cite{videoquality}, the higher score, the better quality. 
(c) The video brightness is measured by \cite{bezryadin2007brightness}, the low variance reflects the more stable in brightness aspect. 
(d) Samples at different brightness, with brightness values in the upper right corner.}
\label{fig:vox_quality}
\end{figure*}

\subsection{Comparison with Video Dataset}

As stated before, VoxCeleb2~\cite{vox2} is one of the largest in-the-wild face video datasets, and it contains massive face videos, that have contributed to the development of many fields \cite{ephrat2018looking,zhu2021deep,gao2021visualvoice,zhou2019talking,chen2019hierarchical,zhu2021arbitrary,zhou2021pose,ji2021audio}. 
However, CelebV-HQ not only contains speech-based videos, we believe that if it is more diverse and of higher quality than the videos in VoxCeleb2~\cite{vox2}, this can be used to further improve the performance of the related models.

To demonstrate the superiority of CelebV-HQ and its ability to better support relevant studies, we compare with VoxCeleb2~\cite{vox2} in terms of the data quality and temporal smoothness. 
For temporal smoothness, we conduct a comprehensive evaluation in brightness, head pose, and motion movement.

\begin{figure*}[t]
\centering
\includegraphics[width=1\textwidth]{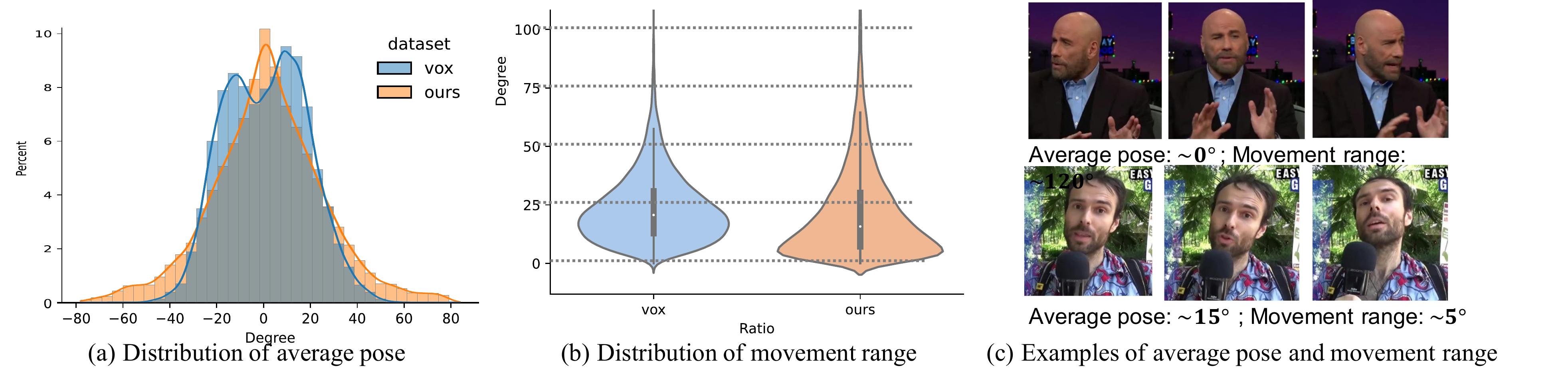}
\caption{\textbf{Distributions of average head pose and movement range.} There is a wide range of head movement in CelebV-HQ, including both stable videos (less than 20$^{\circ}$ of movement) and videos with significant movement (from 75$^{\circ}$ to 100$^{\circ}$).}
\label{fig:vox_pose}
\end{figure*}

\noindent \textbf{Data Quality. } We use BRISQUE~\cite{brisque} as a static quality evaluation metric, which is a non-reference evaluation algorithm. For each video clip, we average the BRISQUE~\cite{brisque} value between frames. 
For the comparison of video quality distributions, we apply the VSFA~\cite{videoquality} measurement, a non-reference evaluation method that scores content dependency and temporal memory effects. 
The distributions are shown in Fig.~\ref{fig:vox_quality}, and CelebV-HQ offers higher quality than VoxCeleb2~\cite{vox2} at both image and video levels.

\noindent \textbf{Brightness Variation. }
We also compare video brightness variance distribution with VoxCeleb2~\cite{vox2}. We first obtain the brightness of each frame and compute the standard deviation in the temporal dimension. 
The brightness is calculated by averaging the pixels and then converting them to ``perceived brightness''~\cite{bezryadin2007brightness}.
The lower variance of brightness indicates more similar luminance within the video clip, \ie, better brightness uniformity. Fig.~\ref{fig:vox_quality}~(c) shows that CelebV-HQ contains more low variance videos, that demonstrates the brightness change during videos in CelebV-HQ is more stable in the temporal dimension.
As shown in Fig.~\ref{fig:vox_quality}~(c), CelebV-HQ contains videos of diverse brightness conditions, to further facilitate the usage of CelebV-HQ, we categorized the videos in terms of brightness, so that users can flexibly choose which samples to use.

\noindent \textbf{Head Pose Distribution.}
The head pose distribution is compared in two aspects: the average head pose of a video and the range of head pose movement. These two are used to show the diversity of head poses across the dataset and within the videos, respectively. As stated before, we leverage~\cite{wu2018look} to detect the head pose in the yaw direction. 
As shown in Fig.~\ref{fig:vox_pose}~(a), CelebV-HQ is more diverse and smoother than VoxCeleb2~\cite{vox2} in the overall distribution. From Fig.~\ref{fig:vox_pose}~(b), we see that we have about 75\% of the data with movements less than 30$^{\circ}$, which means that most of the data are stable, while there are still 25\% of movements between 30$^{\circ}$ and 100$^{\circ}$, indicating the overall distribution is diverse. The illustration of average head pose and  movement range is shown in Fig.~\ref{fig:vox_pose}~(c).

\begin{wrapfigure}{r}{0.5\textwidth}
\vspace{-5mm}
\centering
\includegraphics[width=0.5\textwidth]{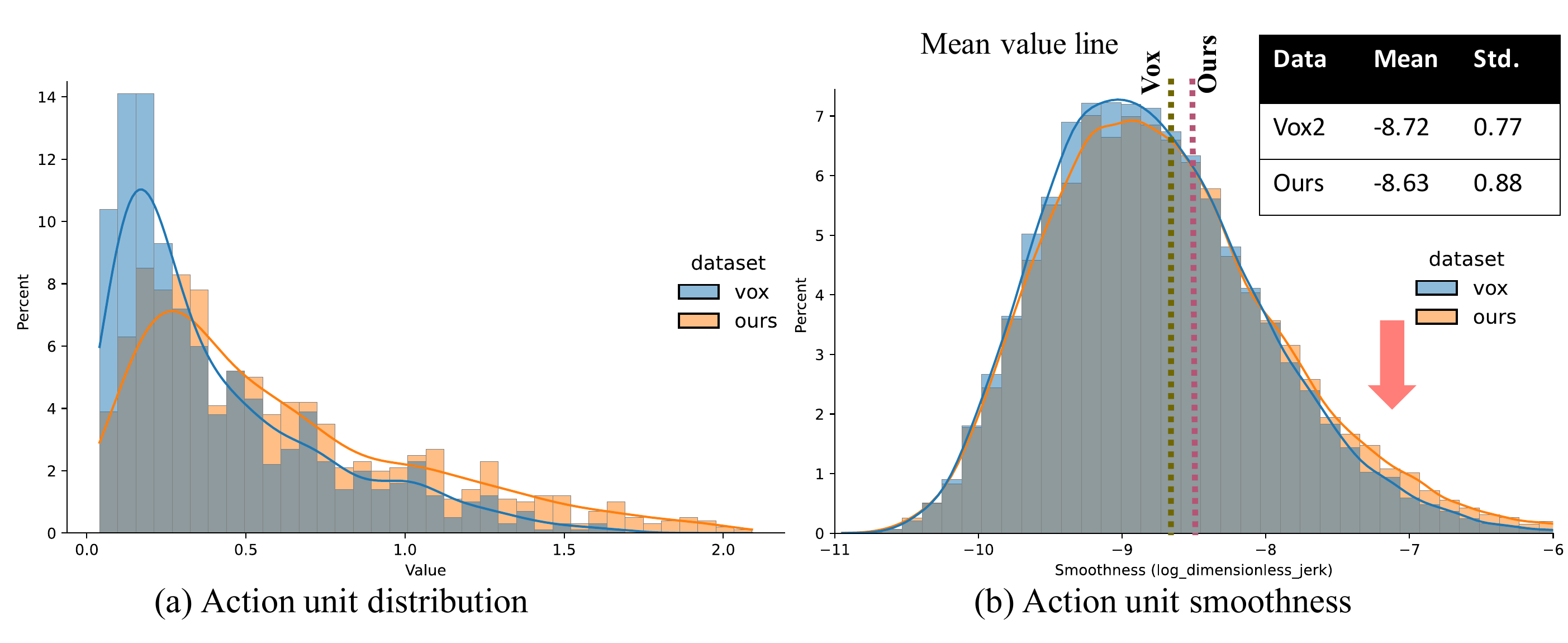}
\caption{\textbf{Distribution and smoothness of action units.} We evaluate the distribution (a) and smoothness (b) of action units.}
\label{fig:au}
\vspace{-5mm}
\end{wrapfigure}

\noindent \textbf{Action Unit Analysis.}
Facial Action Units (AUs) are the basic actions of a muscle or muscle group, and we use~\cite{fan2020fau} to detect AUs. 
The dataset is analyzed in both muscle movement richness and naturalness.
Fig.~\ref{fig:au}~(a) shows that CelebV-HQ is more uniformly distributed over different AU values that represents action strength. The main reason is that videos in VoxCeleb2~\cite{vox2} are mainly talking videos, while CelebV-HQ consists of more types of facial actions. 
Meanwhile, the smoothness is measured by log dimensionless jerk~\cite{ldljerk}. As shown in Fig.~\ref{fig:au}~(b), CelebV-HQ is smoother than VoxCeleb2~\cite{vox2}, as we highlight with the ``Mean value line''. More AU results are presented in Appendix~\ref{sec:additional_au}.

\begin{figure*}[t]
    \centering
    \includegraphics[width=0.90\textwidth]{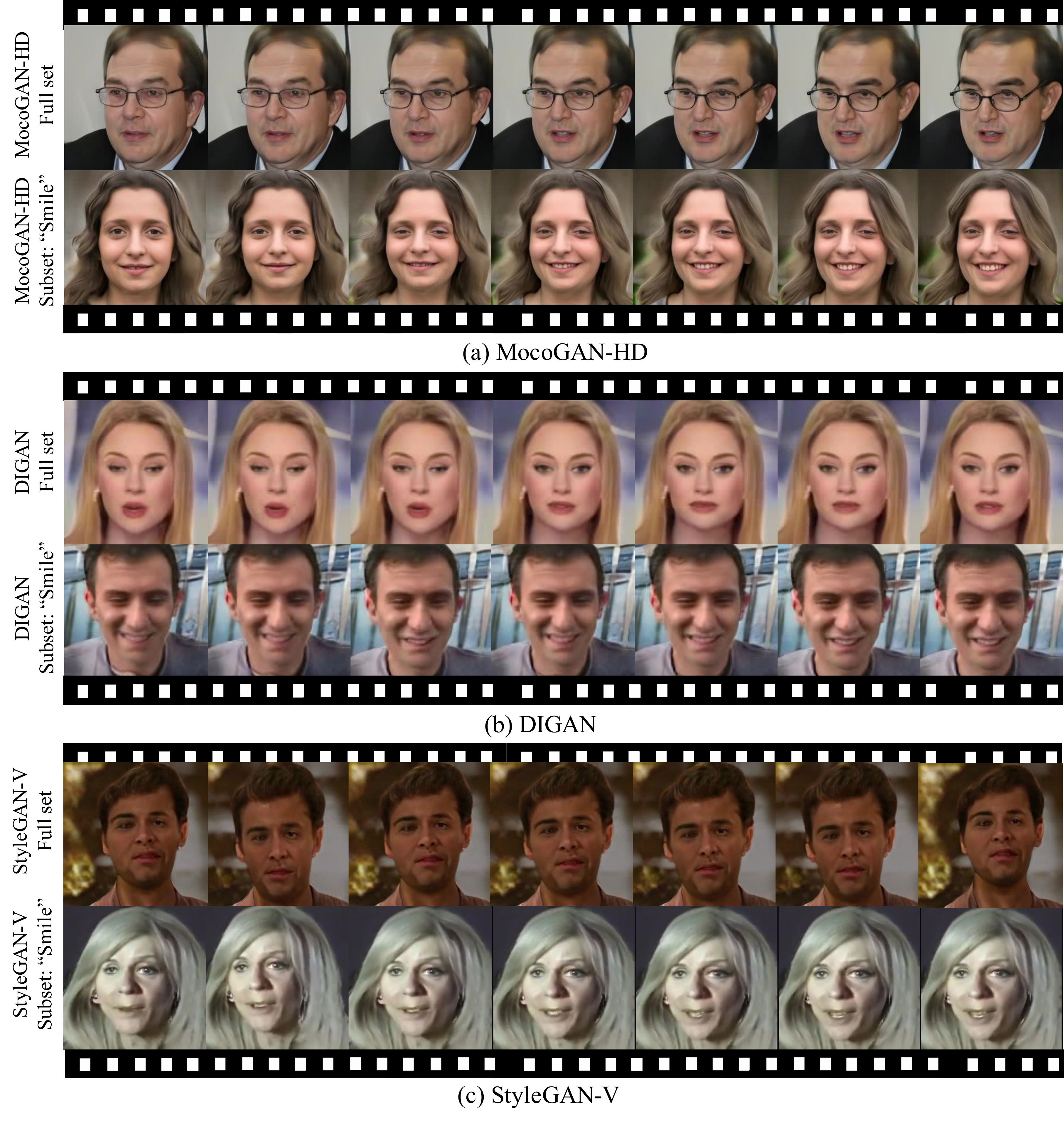}
    \caption{\textbf{Qualitative results of unconditional video generation.} We present ``Full set'' and ``Subset'' settings of MocoGAN-HD~\cite{mocoganhd}, DIGAN~\cite{digan}, and StyleGAN-V~\cite{styleganv} respectively. CelebV-HQ is readily applicable to these unconditional video GANs. }
    \label{fig:unsupervised}
\end{figure*}

\section{Evaluation}
In this section, we describe our experimental setups and the implementation details of baseline methods. We report the results on state-of-the-art baselines in two typical video generation/editing tasks, \ie, unconditional video generation and video facial attribute editing. 
The comprehensive evaluation shows the effectiveness and potential of our proposed CelebV-HQ dataset.

\subsection{Unconditional Video Generation}
\begin{sloppypar}
\noindent
\textbf{Settings.} We employ four unconditional video generation methods, \ie, VideoGPT~\cite{yan2021videogpt}, MoCoGAN-HD~\cite{mocoganhd}, DIGAN~\cite{digan}, and StyleGAN-V~\cite{styleganv}. We chose these methods based on their performance and code availability.
Furthermore, since CelebV-HQ contains action annotations, the models are evaluated under two settings, \ie, the full set of data and the subsets split by different action attributes, \eg, smile. 
We strictly followed the original authors' setting. The videos generated by MoCoGAN-HD~\cite{mocoganhd} and StyleGAN-V~\cite{styleganv} are 256$\times$256 resolution, and the one generated by VideoGPT~\cite{yan2021videogpt} and DIGAN~\cite{digan} are 128$\times$128 resolution.
To evaluate the model performance, we leverage FVD~\cite{fvd} and FID~\cite{fid} to access video quality and image quality, respectively. Please refer to Appendix~\ref{sec:setting_details} for more details.
\end{sloppypar}

\begin{table*}[t]
\centering
\vspace{-2mm}
\caption{\textbf{Quantitative results of unconditional video generation.} We evaluate VideoGPT~\cite{yan2021videogpt}, MoCoGAN-HD~\cite{mocoganhd}, DIGAN~\cite{digan}, and StyleGAN-V~\cite{styleganv} on different datasets and report the FVD and FID scores. ``$\downarrow$'' means a lower value is better.}
\resizebox{1\textwidth}{!}{
\setlength{\tabcolsep}{0.0190\textwidth}{
\begin{tabular}{l|cc|cc|cc|cc}
\specialrule{.12em}{.1em}{.1em} \textbf{}  & \multicolumn{2}{c|}{FaceForensics~\cite{faceforensics}}                   & \multicolumn{2}{c|}{Vox~\cite{vox17}}   & \multicolumn{2}{c|}{MEAD~\cite{mead}} & \multicolumn{2}{c}{CelebV-HQ} \\
                            & FVD  ($\downarrow$)   & FID ($\downarrow$)     & FVD ($\downarrow$)        & FID ($\downarrow$)     & FVD  ($\downarrow$)        &  FID  ($\downarrow$)       & FVD  ($\downarrow$)      & FID ($\downarrow$)   \\ \hline
VideoGPT~\cite{yan2021videogpt}  & 185.90  &  38.19   & 187.95     & 65.18   & 233.12      &  75.32      & 177.89    & 52.95 \\
MoCoGAN-HD~\cite{mocoganhd} & 111.80  & \textbf{7.12}   & 314.68     & \textbf{55.98}   & 245.63      &  32.54      & 212.41    & 21.55  \\
DIGAN~\cite{digan}          & 62.50   & 19.10   & 201.21     & 72.21   & 165.90      &  43.31      & 72.98    & 19.39 \\
StyleGAN-V~\cite{styleganv} & \textbf{47.41}   & 9.45    & \textbf{112.46}    & 60.44   & \textbf{93.89}       &  \textbf{31.15}      &  \textbf{69.17} & \textbf{17.95} \\
\specialrule{.12em}{.1em}{.1em} 
\end{tabular}
}
}
\label{tbl:unsupervise}
\vspace{-2mm}
\end{table*}


\begin{figure*}[t]
    \centering
    \vspace{-1mm}
    \includegraphics[width=0.90\textwidth]{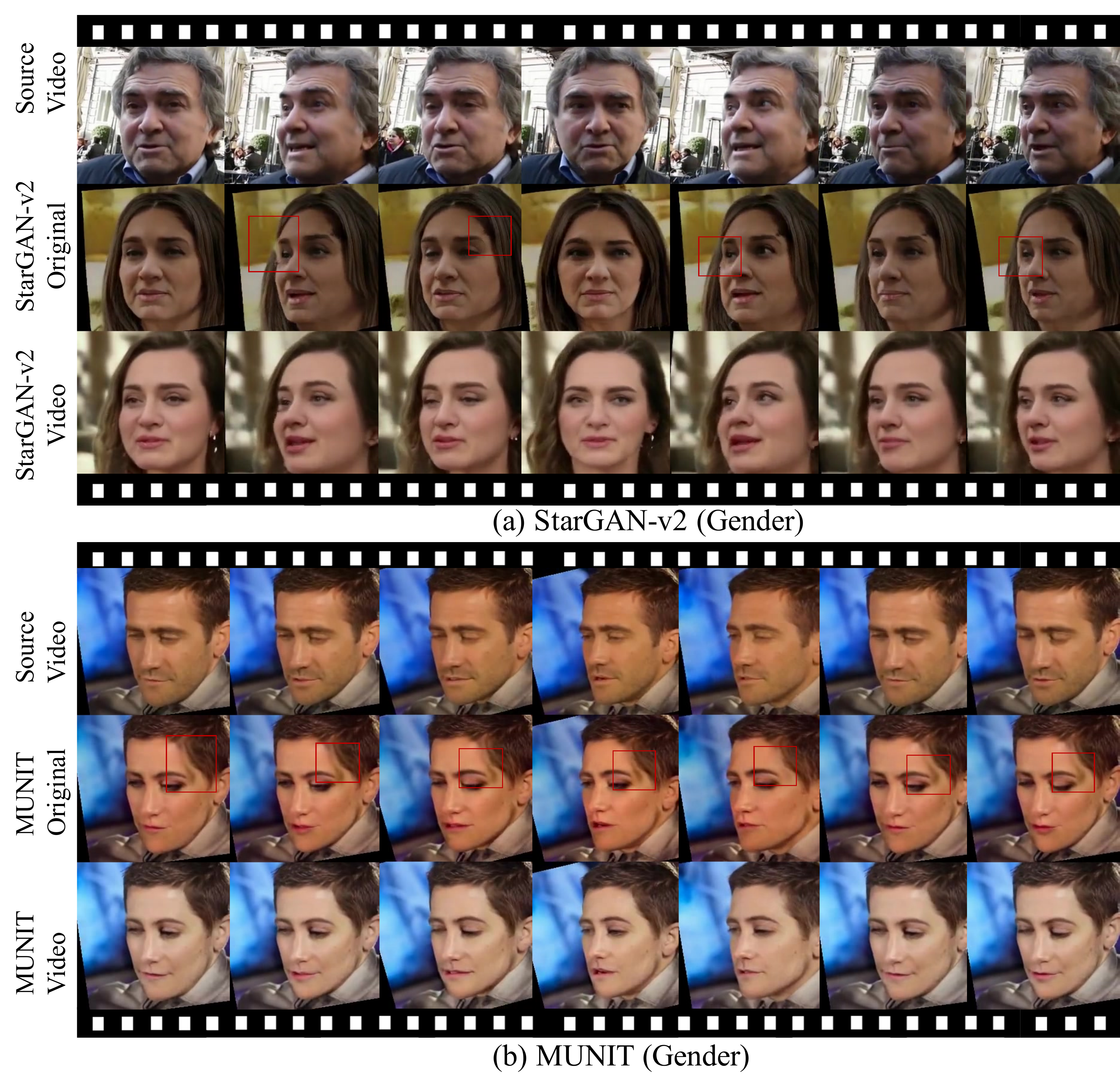}
    \vspace{-3mm}
    \caption{\textbf{Qualitative results of video facial attribute editing.} Results of ``Original'' tend to have a jittering in the hair area, while results of ``Video'' are more stable. }
    \label{fig:i2i}
    
\end{figure*}

\noindent
\textbf{Results.}
As shown in Fig.~\ref{fig:unsupervised}, we first focus on temporal consistency. MocoGAN-HD~\cite{mocoganhd}, DIGAN~\cite{digan}, and StyleGAN-V\cite{styleganv} can generate consistent videos trained on CelebV-HQ.
Besides, all methods can successfully produce the desire actions when trained on different subsets of CelebV-HQ with specific attributes. 
Satisfactory results achieved on the these state-of-the-art methods, demonstrating the effectiveness of CelebV-HQ. 

\noindent
\textbf{Benchmark.}
We construct a benchmark of unconditional video generation task, for four currently prevalent models (VideoGPT~\cite{yan2021videogpt}, MoCoGAN-HD~\cite{mocoganhd}, DIGAN~\cite{digan}, and StyleGAN-V~\cite{styleganv}) on $4$ face video datasets (FaceForensics~\cite{faceforensics}, Vox~\cite{vox17}, MEAD~\cite{mead} and CelebV-HQ). 
The benchmark is presented in Table~\ref{tbl:unsupervise}. 
Firstly, it can be observed that the ranking achieved by CelebV-HQ is similar to other prevalent datasets within different methods, which indicates the effectiveness of CelebV-HQ. 
In addition, the current video generation models~\cite{yan2021videogpt,mocoganhd,digan,styleganv} obtained good FVD/FID metrics compared to the Vox~\cite{vox17} dataset with similar data size. This illustrates that CelebV-HQ further exploits the potential of the current work, allowing it to generate more diverse and higher quality results.
However, as a challenging real-world dataset, CelebV-HQ still has a considerable room for community to make improvement.

\begin{table}[t]
\vspace{-2mm}
\caption{\textbf{Quantitative results of video facial attribute editing.} We evaluate two video facial editing baselines. The ``Video'' version achieves lower FVD scores and comparable FID performance than ``Original''. ``$\downarrow$'' means a lower value is better.}
\centering
\resizebox{0.7\textwidth}{!}{
\begin{tabular}{l|cccc|cc}
\specialrule{.12em}{.1em}{.1em}

           & \multicolumn{4}{c|}{StarGAN-v2 (Gender)}                                                       & \multicolumn{2}{c}{MUNIT (Gender)}                         \\ \cline{2-7} 
    Metrics  & \multicolumn{2}{c}{Original}                     & \multicolumn{2}{c|}{Video}                     & \multirow{2}{*}{Original} & \multirow{2}{*}{Video}            \\ \cline{2-5}
           & Reference                     & Label          & \multicolumn{1}{c}{Reference} & Label         &                         &                                  \\ \hline
FVD ($\downarrow$)  & 284.80                        & 258.36  & \textcolor{blue}{\textbf{262.01}   }                     & \textcolor{blue}{\textbf{189.40}} & 219.96                  & \textcolor{blue}{\textbf{211.45} }\\
FID ($\downarrow$) & \textbf{80.61}                         & 65.70    & 82.99                         & \textbf{55.73}   & 58.58                   & \textbf{57.01}                            \\  
\specialrule{.12em}{.1em}{.1em}

\end{tabular}
}
\label{tbl:i2i}
\vspace{-5mm}
\end{table}

\subsection{Video Facial Attribute Editing}
\noindent
\textbf{Settings.}
We employ two representative facial editing baselines, \ie, StarGAN-v2~\cite{starganv2} and MUNIT~\cite{munit}, to explore the potential of CelebV-HQ on video facial attribute editing task.
The canonical StarGAN-v2~\cite{starganv2} and MUNIT~\cite{munit} are desiged for static image data. To evaluate the effectiveness of the proposed video dataset, we modify these models by simply adding a vanilla temporal constraint, \ie, estimating the optical flows for $i$-th frame and $(i+t)$-th frame in different domains by LiteFlowNet~\cite{hui20liteflownet3} and enforcing L2 Loss between the flows. Other losses the original authors proposed remain unchanged.
To demonstrate the practical value of our dataset, we select a common used appearance attribute, \ie, ``Gender'', for different baselines. 


\noindent
\textbf{Results.}
The baseline methods achieve good results when editing the Gender attribute. The main difference lies in the temporal consistency. 
%
%
In Fig.~\ref{fig:i2i}, we observe that the results generated by the original image models are sometimes unstable in the hair area. For instance, StarGAN-v2 suffers hair shape inconsistencies, and MUNIT produces jittering color block defects in hair generation. 
As reported in Table~\ref{tbl:i2i}, the ``Video'' version outperform the ``Original'' one with respect to the FVD metric in all cases (highlighted in \textcolor{blue}{blue}).
Furthermore, the ``Video'' version obtains comparable FID results as ``Original''. 
%
These results indicate that a simple modification using the temporal cues in video can bring performance enhancement, which demonstrate the effectiveness of the proposed dataset in video facial editing tasks. Please refer to the Appendix~\ref{sec:additional_editing_results} for more attributes results.
%

\section{Discussion}


\subsection{Empirical Insights}
Some empirical insights are drawn during the construction of CelebV-HQ and the baseline benchmarking. 

1) We observe a trend in the growing demand for video facial editing due to the prevalence of short videos, \eg, TikTok~\cite{tiktok} and Snapchat~\cite{snapchat}. However, as we stated before, current applications are mainly based on static images~\cite{faceapp,snapchat}. 
Therefore, the research on transforming face editing from images to videos would be an emerging direction in the future. 

2) An effective video alignment strategy is important for coherent video generation.
In most image generation studies, faces are usually aligned by key points, which may be jittery if applied directly in video generation~\cite{bilayermodel}. On the other hand, the generation quality might degrade if faces in the video are not aligned. 
This suggests a new method that can simultaneously retain temporal information and align the face may improve the temporal consistency of generated videos.


\subsection{Future Work}
\begin{sloppypar}
    
Finally, we envision the research areas that may benefit from CelebV-HQ. 

%

\noindent
\textbf{Video Generation/Editing.}
CelebV-HQ provides the possibility of improving Video Generation/Editing, such as unconditional face generation~\cite{gan,dcgan,stylegan,stylegan2,mocoganhd,digan,styleganv}, text-to-video generation~\cite{li2018video,nuwa,videodiffusion,cogvideo,text2live}, video facial attributes editing~\cite{starganv2,munit,interfacegan,e4e,xu2022transeditor}, face reenactment~\cite{wayne2018reenactgan,fomm,bilayermodel,wang2021facevid2vid}, and face swapping~\cite{li2019faceshifter,zhu2020aot,Gao_2021_CVPR,nirkin2019fsgan,xu2022mobilefaceswap}. 
These tasks rely heavily on the scale and quality of the dataset for generalization. 
Moreover, some video generation methods still leverage the frame-level quality while neglecting the temporal information~\cite{siarohin2021motion,wang2021one,Zhou_2021_CVPR}. Nevertheless, temporal modeling is essential for generating smooth and realistic videos, which deserves further investigation. 
Since CelebV-HQ also contains rich annotations of facial attributes, this would allow researcher to go deeper when using these information, \eg, synthesize text description with templates and learning disentanglement of facial attributes.

\noindent
\textbf{Neural Rendering.} CelebV-HQ has great potential for applications in Neural Rendering. Current tasks, such as novel view synthesis \cite{headnerf,stylenerf,or2021stylesdf,DBLP:journals/corr/abs-2204-11798,DBLP:journals/corr/abs-2201-07786} and 3d generation~\cite{chan2021pigan,or2021stylesdf,eg3d,guo2021adnerf,Gafni_2021_CVPR,chen2022sem2nerf}, are trained on in-the-wild image datasets~\cite{celebahq,stylegan} which lacks facial dynamics to provide natural geometries.
CelebV-HQ, as a high-quality and large-scale video dataset, provides natural facial dynamics and diverse 3D geometries. These features on video modality could not only be further exploited to improve the quality of current models, but also stimulate the emerging of several budding topics, such as Dynamic NeRF~\cite{pumarola2021d} and Animatable NeRF~\cite{peng2021animatable}.

\noindent
\textbf{Face Analysis.} Face Analysis tasks, such as Attribute Recognition~\cite{zhong2016face,ding2018deep,fairface}, Action Recognition~\cite{wang2013action,jegham2020vision}, Emotion Recognition~\cite{dzedzickis2020human,lee2019context}, Forgery Detection~\cite{li2020face,haliassos2021lips,zhu2021face}, and Multi-modal Recognition~\cite{zhang2020emotion,munro2020multi}. These tasks usually require the dataset to have diverse attribute coverage and natural distribution. 
CelebV-HQ not only meets these requirements, but also could helps to transfer previous image tasks, such as attribute recognition, to the video version by learning spatio-temporal representations. 
\end{sloppypar}

\section{Conclusion}

In this paper, we propose a large-scale, high-quality, and diverse video dataset with rich facial attributes, called CelebV-HQ. CelebV-HQ contains $35,666$ video clips involving $15,653$ identities, accompanied by $40$ appearance attributes, $35$ action attributes, and $8$ emotion attributes.
Through extensive statistical analysis of the dataset, we show the rich diversity of CelebV-HQ in terms of age, ethnicity, brightness, motion smoothness, pose diversity, data quality, \textit{etc}.
The effectiveness and future potential of CelebV-HQ are also demonstrated via the unconditional video generation and video facial attribute editing tasks. 
Finally, we provide an outlook on the future prospects of CelebV-HQ, which we believe can bring new opportunities and challenges to the academic community. In the future, we are going to maintain a continued evolution of CelebV-HQ, including the scale, quality and annotations.

\noindent
\textbf{Acknowledgement.} This work is partly supported by Shanghai AI Laboratory and SenseTime Research. It is also supported by NTU NAP, MOE AcRF Tier 1 (2021-T1-001-088), and under the RIE2020 Industry Alignment Fund – Industry Collaboration Projects (IAF-ICP) Funding Initiative, as well as cash and in-kind contribution from the industry partner(s).

{\small
\bibliographystyle{ieee_fullname}
\bibliography{egbib}
}


\clearpage
\appendix

\noindent
\textbf{\LARGE Appendix}

\setcounter{table}{0}
\renewcommand{\thetable}{A\arabic{table}}
\setcounter{figure}{0}
\renewcommand{\thefigure}{A\arabic{figure}}

\section{Data Pre-processing}
\label{appsec:video_preprocess}
\begin{figure*}[h]
\centering
\includegraphics[width=1\textwidth]{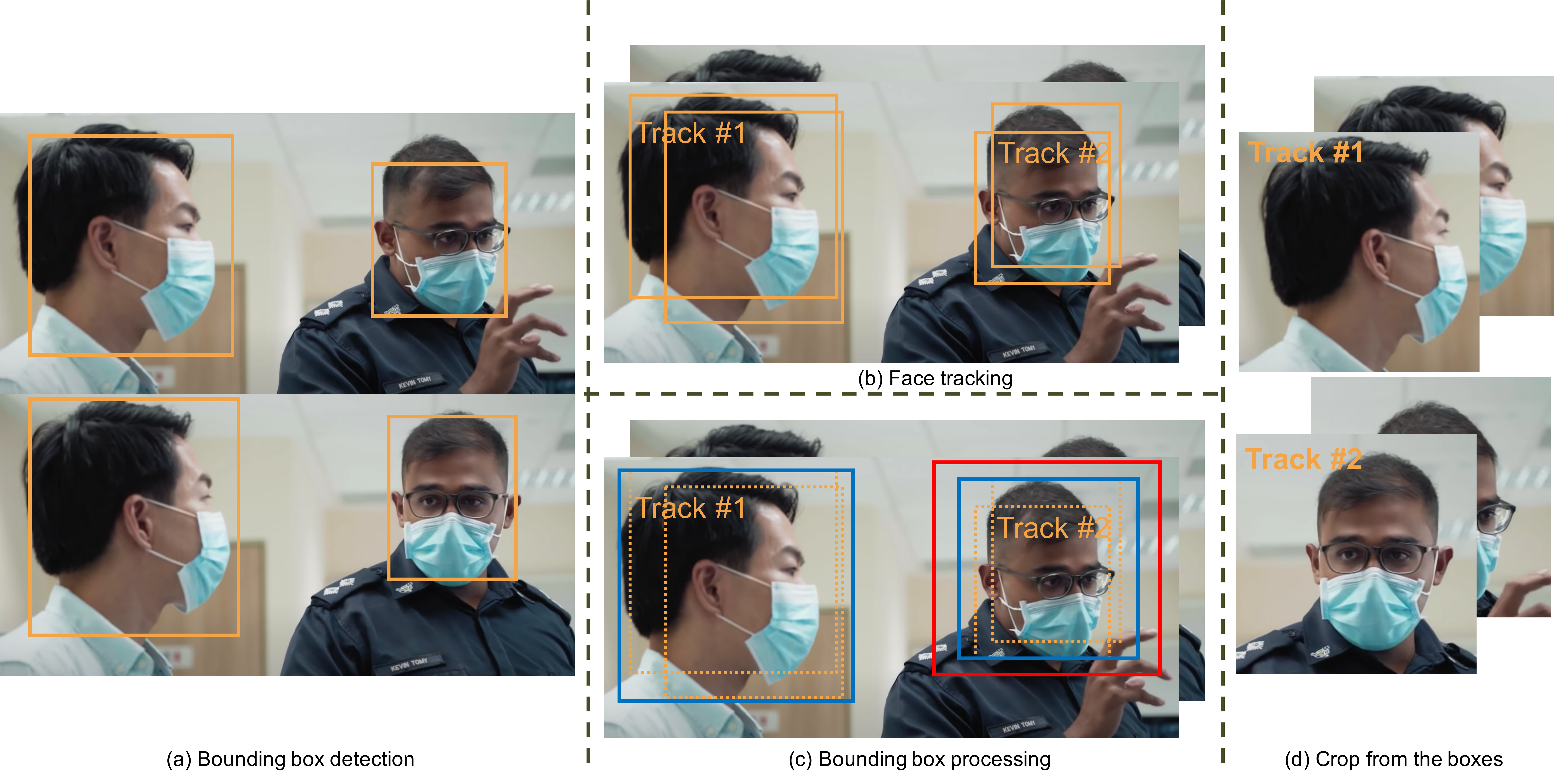}
\caption{\textbf{Pipeline of data pre-process} (a) We start from the bounding box detection for each frame. (b) A tracking framework~\cite{sort} is introduced to track different identities. (c) Given bounding box sequences (\textcolor{orange}{dotted orange boxes}), we calculate their minimum bounding rectangles (\textcolor{blue}{blue box}). If bounding rectangles smaller than 512$\times$512, we expand it to this size (\textcolor{red}{red box}). (d) Finally, the videos are cropped using the bounding rectangles (blue/red boxes).  }
\label{fig:data_preprocess}
\end{figure*}

We provide a illustration of data pre-processing as shown in Fig.~\ref{fig:data_preprocess}. The detailed description is stated in the Sec.~\ref{sec:video_preprocess}. 

\begin{figure*}[t]
\centering
\includegraphics[width=1\textwidth]{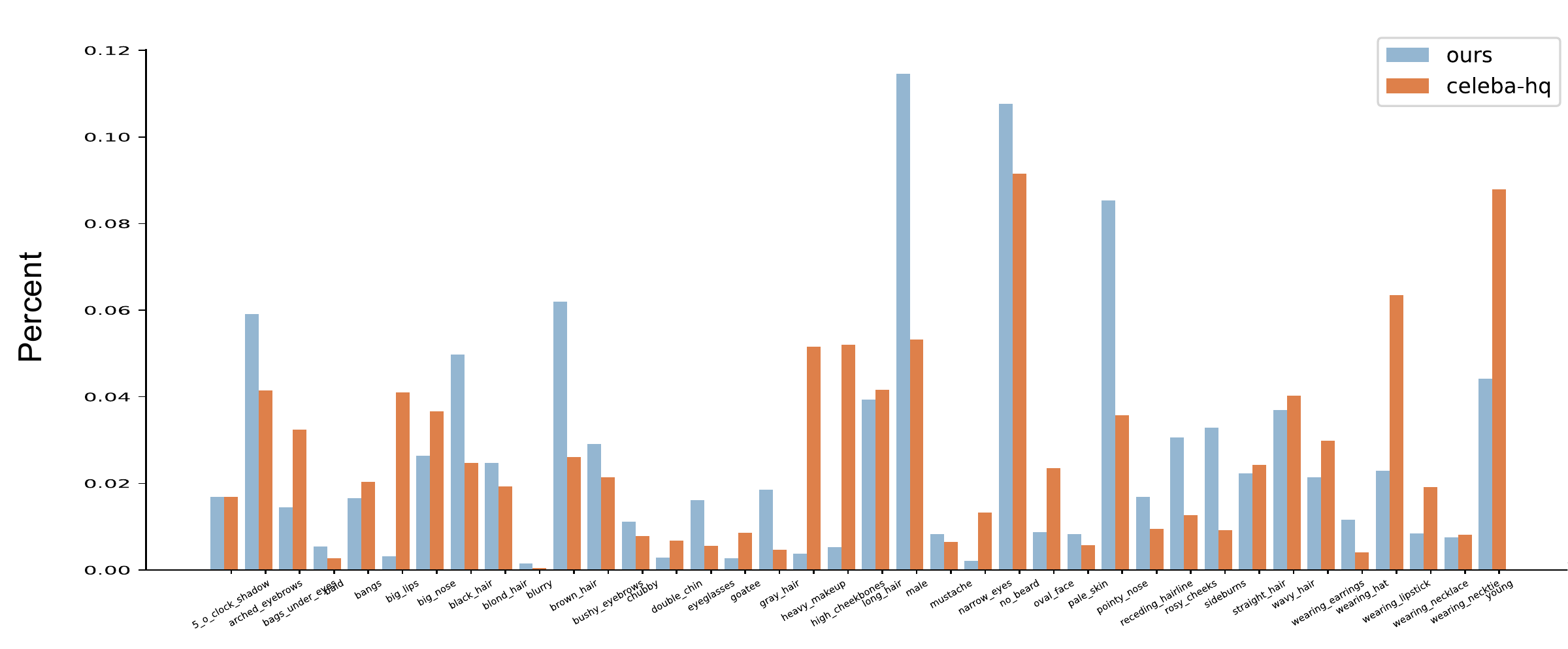}
\caption{\textbf{Comparison of appearance attribute statistics with CelebA-HQ~\cite{celebahq}.} Please zoom in for more details.}
\label{fig:attr_comp_celebahq}
\vspace{-4mm}
\end{figure*}

\begin{figure}[t]
\centering
\includegraphics[width=1\textwidth]{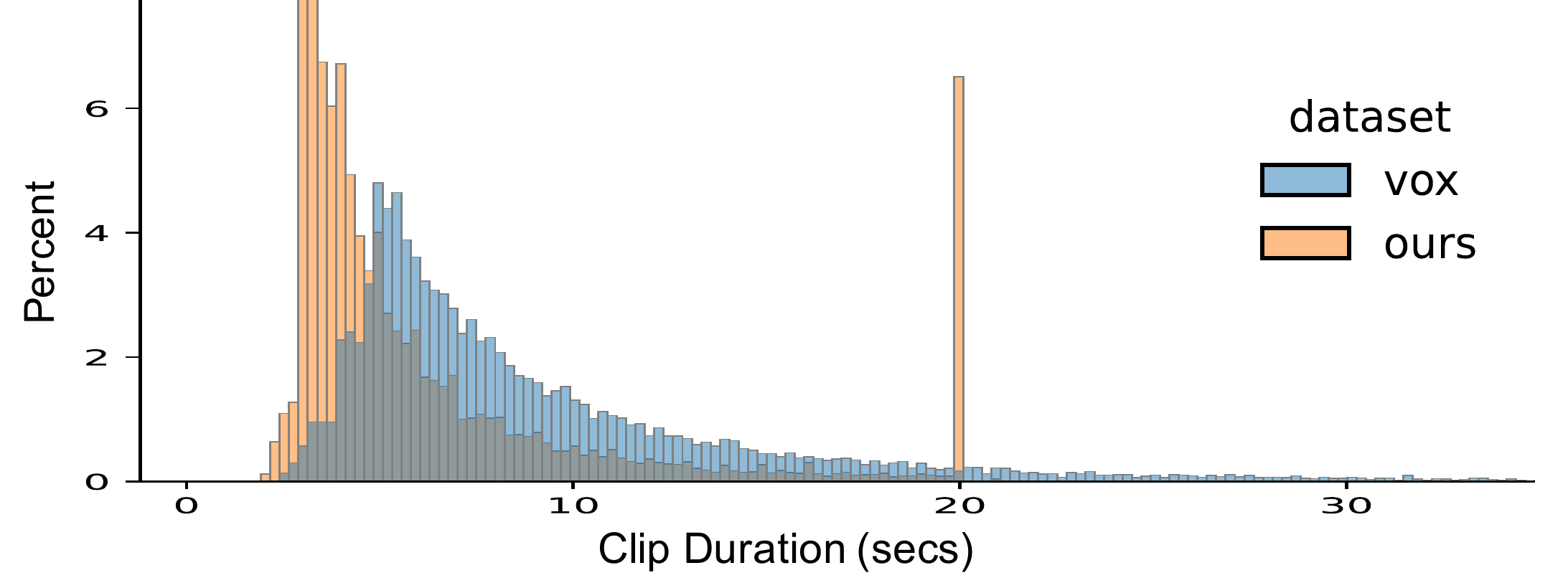}
\caption{\textbf{Comparison of clip duration statistics with Vox~\cite{vox17}.} }
\label{fig:dua_comp_vox}
\vspace{-4mm}
\end{figure}

\section{Additional Statistic Comparisons}
\subsection{Comparison of appearance attribute statistics with CelebA-HQ}
As shown in Fig.~\ref{fig:attr_comp_celebahq}, CelebV-HQ has a similar distribution to CelebA-HQ, and the distribution of most appearance attributes is close to that of CelebA-HQ. This indicates that there is no significant deviation in the distribution of CelebV-HQ.
 
\subsection{Comparison of clip duration statistics with Vox}
As reported in Fig.~\ref{fig:dua_comp_vox}, the clip time distribution is shorter compared to Vox~\cite{vox17} for ensuring video consistency and annotation accuracy. Also, the videos in CelebV-HQ are all less than 20s, this is because we truncate all the videos at 20s to avoid the attributes changing in the long video.

\subsection{Additional AUs Distribution}
\label{sec:additional_au}
\begin{figure*}[t]
\centering
\includegraphics[width=1\textwidth]{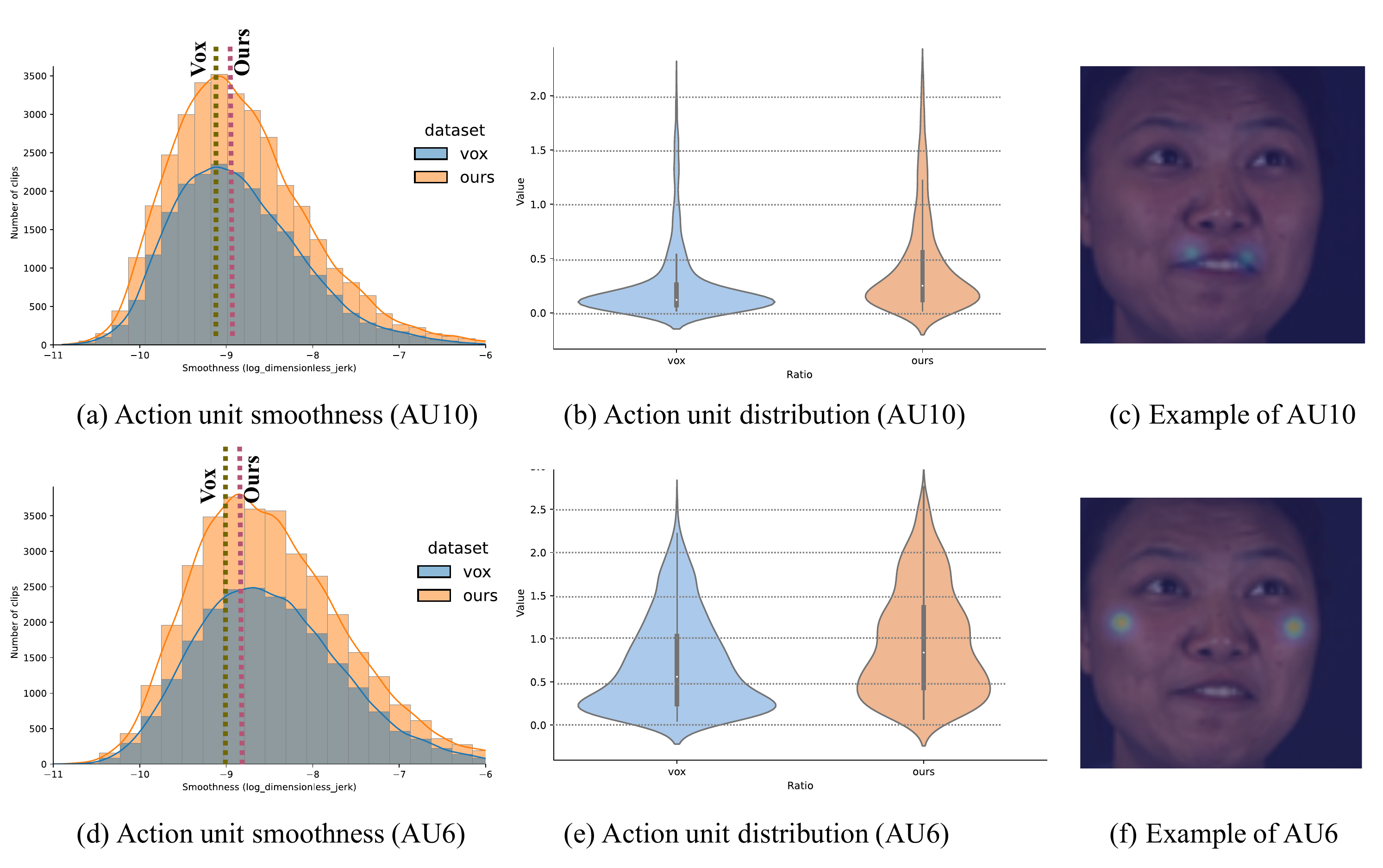}
\caption{\textbf{Distributions of different AUs.} }
\label{fig:AU_more}
\end{figure*}

We provide additional action units (AUs) distributions as shown in Fig.\ref{fig:AU_more}. Fig.~\ref{fig:AU_more}~(c) and (f) show the locations represented by the different AUs. 
In Fig.~\ref{fig:AU_more}~(a) and (d), we can see that the action of CelebV-HQ is smoother than VoxCeleb2~\cite{vox2}. Meanwhile, Fig.~\ref{fig:AU_more}~(b) and (e) suggest that CelebV-HQ is more evenly distributed at different AU values.

\section{Additional Experiments}
\begin{table}[t]
\vspace{-2mm}
\caption{\textbf{Quantitative results of video facial attribute editing.} We evaluate two video facial editing baselines. The ``Video'' version achieves lower FVD scores and comparable FID performance than ``Original''. ``$\downarrow$'' means a lower value is better.}
\centering
\resizebox{0.8\textwidth}{!}{
\begin{tabular}{l|cccc|cc}
\specialrule{.12em}{.1em}{.1em} 
           & \multicolumn{4}{c|}{\textbf{StarGAN-v2 (Brown Hair)}}                                                   & \multicolumn{2}{c}{\textbf{MUNIT (Eyeglasses)}}                     \\ \cline{2-7} 
    \textbf{Metrics}       & \multicolumn{2}{c}{Original}                     & \multicolumn{2}{c|}{Video}                     & \multirow{2}{*}{Original} & \multirow{2}{*}{Video}            \\ \cline{2-5}
           & \multicolumn{1}{c}{Reference} & Label          & \multicolumn{1}{c}{Reference} & Label         &                         &                                  \\ \hline
FVD ($\downarrow$)& 323.71                        & 244.58 & \textcolor{blue}{\textbf{295.74}}                        & \textcolor{blue}{\textbf{232.63}} & 204.12                  & \textcolor{blue}{\textbf{158.87}} \\
FID ($\downarrow$)& \textbf{77.26}                         & \textbf{64.82}   & 89.07                         & 69.68   & \textbf{30.65}                   & 31.23 \\                           \specialrule{.12em}{.1em}{.1em} 
\end{tabular}
}
\label{tbl:i2i_supp}
\end{table}

\subsection{FVD/FID Setting Details}
\label{sec:setting_details}
We leverage FID\footnote{\href{https://github.com/mseitzer/pytorch-fid}{https://github.com/mseitzer/pytorch-fid}}~\cite{fid} and FVD\footnote{\href{https://github.com/sihyun-yu/digan/tree/master/src/metrics}{https://github.com/sihyun-yu/digan/tree/master/src/metrics}}~\cite{fvd} to assess the image and video quality of the video generation and editing models. As both metrics are sensitive to the amount of data in the test set, we first select 2048 videos randomly as our test set. All videos in the test set are used as the ``real'' part in the metric experiments. 
For the unconditional generation, we also randomly generate 2048 videos as the ``fake'' part.
For the editing of video facial attribution, we generate corresponding fake results for each real video, yielding 2048 fake videos as well.
To provide enough images for FID testing, we sample 4 frames from each video. In total, we have 8192 images for the real data and fake data respectively. 
For the FVD, we use all the real and generated videos.

\begin{figure*}[t]
\centering
\includegraphics[width=1\textwidth]{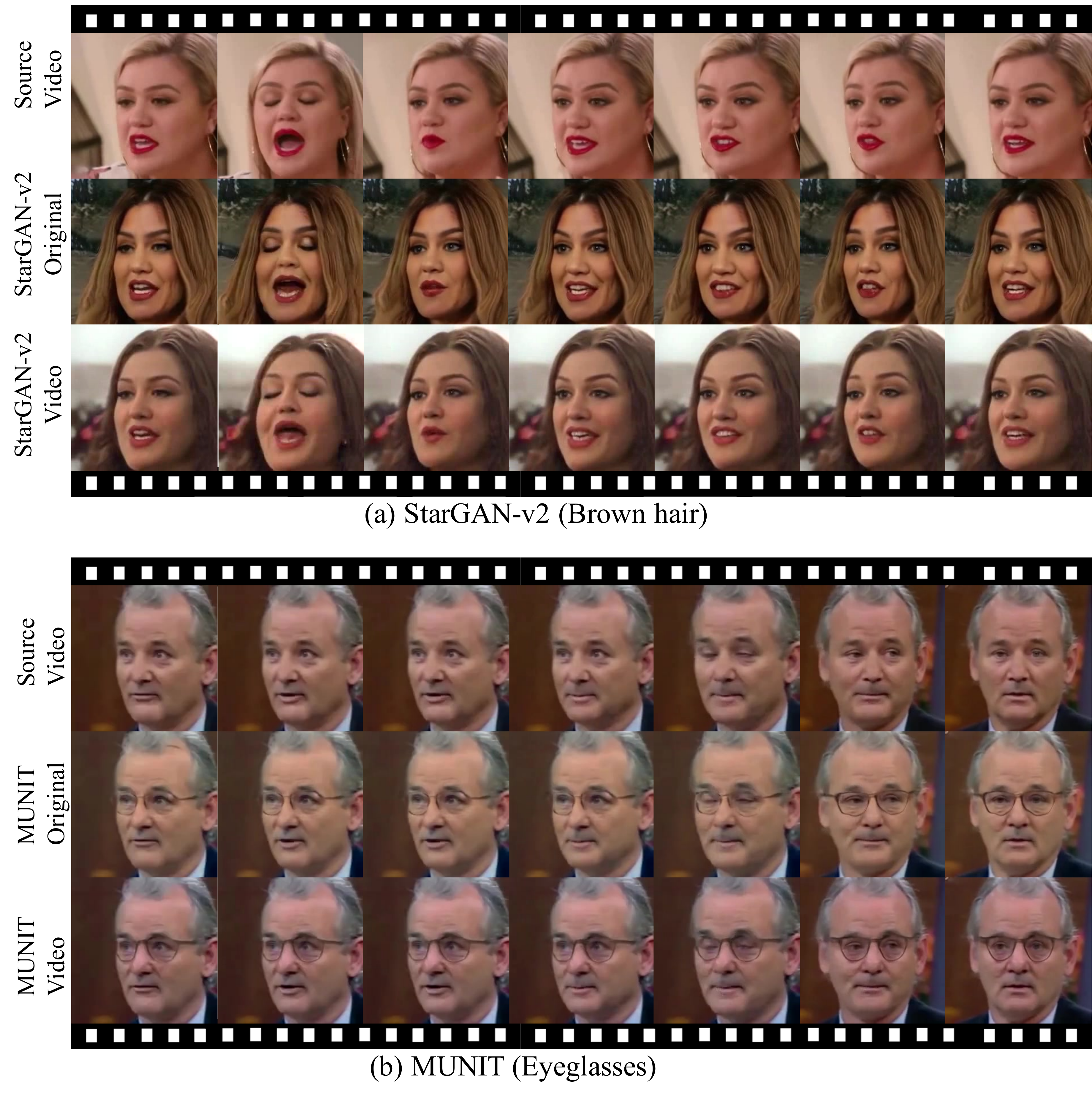}
\caption{\textbf{Qualitative results of video facial attribute editing.} In (a), we edit the attribute \emph{brown hair} with StarGAN-v2~\cite{starganv2}. In (b), we edit the attribute \emph{eyeglasses} with MUNIT~\cite{munit}. }
\label{fig:i2i_supp}
\vspace{-1mm}
\end{figure*}

\subsection{Additional Video Facial Attribute Editing Results}
\label{sec:additional_editing_results}
To demonstrate the practical value of our dataset for facial attributes editing in low-level appearance attribute. We additional select ``Brown Hair'' attributes for StarGAN-v2~\cite{starganv2}, as well as ``Eyeglasses'' attributes for MUNIT~\cite{munit}. 
The additional results are reported in Table~\ref{tbl:i2i_supp} and Fig.~\ref{fig:i2i_supp}. By simply adding a temporal regularization term, we improve the results of StarGAN-v2~\cite{starganv2} and MUNIT~\cite{munit} in terms of realism and coherence. Note that the temporal regularization is enabled by CelebV-HQ which contains rich annotations and facial dynamics.

\begin{table}[t]
\caption{Quantitative results of video facial attribute editing.}
\centering
\resizebox{1\textwidth}{!}{

\begin{tabular}{l|cccc|cccc}
\specialrule{.12em}{.1em}{.1em} 
\textbf{} & \multicolumn{4}{c|}{\textbf{StarGAN-v2 (Gender)}}                     & \multicolumn{4}{c}{\textbf{StarGAN-v2 (Brown hair)}}                 \\ \cline{2-9} 
\textbf{Metrics}   & \multicolumn{2}{c}{Vox-labeled} & \multicolumn{2}{c|}{CelebV-HQ (Ours)} & \multicolumn{2}{c}{Vox-labeled} & \multicolumn{2}{c}{CelebV-HQ (Ours)} \\ \cline{2-9} 
          & Reference        & Label        & Reference     & Label      & Reference        & Label        & Reference     & Label     \\ \hline
FVD ($\downarrow$)      & 568.79           & 629.09       & \textbf{262.01}        & \textbf{189.04}     & 542.88           & 500.77      & \textbf{295.74}        & \textbf{232.63}    \\
FID ($\downarrow$)      &      104.00      & 85.14        & \textbf{82.99}         & \textbf{55.73}      & 99.57           & 131.18        & \textbf{89.07}         & \textbf{69.68} \\   
\specialrule{.12em}{.1em}{.1em} 
\end{tabular}

}
\label{tbl:i2i_labeled}
\end{table}

\subsection{Experiment on labeled Vox}
We labeled Vox dataset~\cite{vox17} using an open-source algorithm\footnote{\href{https://github.com/ewrfcas/face_attribute_classification_pytorch}{\text{https://github.com/ewrfcas/face\_attribute\_classification\_pytorch}}}. As reported in Table~\ref{tbl:i2i_labeled}, models trained on CelebV-HQ yields better performance. Experiment verified algorithmically labeling existing dataset is not suitable substitutes for CelebV-HQ.

\section{Complete Attributes List}
\label{sec:attribute_list}
\begin{table*}[t]
\centering
\caption{ \textbf{Complete attribute list.} CelebV-HQ contains 83 annotations, including 40 appearance attributes, 35 action attributes, and 8 emotion attributes.}
\resizebox{1\textwidth}{!}{
\begin{tabular}{cccccccc}
\hline
\multicolumn{8}{c}{\textbf{(a) Appearance Attribute}} \\ \hline
\multicolumn{1}{c|}{blurry}                                                & \multicolumn{1}{c|}{male}                                                      & \multicolumn{1}{c|}{young}                                                      & \multicolumn{1}{c|}{chubby}                                                      & \multicolumn{1}{c|}{\begin{tabular}[c]{@{}l@{}}pale\_skin\end{tabular}}        & \multicolumn{1}{c|}{\begin{tabular}[c]{@{}l@{}}rosy\_cheeks\end{tabular}}        & \multicolumn{1}{c|}{\begin{tabular}[c]{@{}l@{}}oval\_face\end{tabular}}       & \begin{tabular}[c]{@{}l@{}}receding\\ hairline\end{tabular} \\ \hline
\multicolumn{1}{c|}{bald}                                                  & \multicolumn{1}{c|}{bangs}                                                     & \multicolumn{1}{c|}{\begin{tabular}[c]{@{}l@{}}black\_hair\end{tabular}}       & \multicolumn{1}{c|}{\begin{tabular}[c]{@{}l@{}}blond\_hair\end{tabular}}       & \multicolumn{1}{c|}{\begin{tabular}[c]{@{}l@{}}gray\_hair\end{tabular}}        & \multicolumn{1}{c|}{\begin{tabular}[c]{@{}l@{}}brown\_hair\end{tabular}}         & \multicolumn{1}{c|}{\begin{tabular}[c]{@{}l@{}}straight\\ hair\end{tabular}}   & \begin{tabular}[c]{@{}l@{}}wavy\_hair\end{tabular}         \\ \hline
\multicolumn{1}{c|}{long\_hair}                                            & \multicolumn{1}{c|}{\begin{tabular}[c]{@{}l@{}}arched\\ eyebrows\end{tabular}} & \multicolumn{1}{c|}{\begin{tabular}[c]{@{}l@{}}bushy\\ eyebrows\end{tabular}}   & \multicolumn{1}{c|}{\begin{tabular}[c]{@{}l@{}}bags\_under\_eyes\end{tabular}} & \multicolumn{1}{c|}{eyeglasses}                                                 & \multicolumn{1}{c|}{sunglasses}                                                   & \multicolumn{1}{c|}{\begin{tabular}[c]{@{}l@{}}narrow\_eyes\end{tabular}}     & \begin{tabular}[c]{@{}l@{}}big\_nose\end{tabular}          \\ \hline
\multicolumn{1}{c|}{\begin{tabular}[c]{@{}l@{}}pointy\_nose\end{tabular}} & \multicolumn{1}{c|}{\begin{tabular}[c]{@{}l@{}}high\\ cheekbones\end{tabular}} & \multicolumn{1}{c|}{\begin{tabular}[c]{@{}l@{}}big\_lips\end{tabular}}         & \multicolumn{1}{c|}{\begin{tabular}[c]{@{}l@{}}double\_chin\end{tabular}}       & \multicolumn{1}{c|}{\begin{tabular}[c]{@{}l@{}}no\_beard\end{tabular}}         & \multicolumn{1}{c|}{\begin{tabular}[c]{@{}l@{}}5\_o\_clock\\ shadow\end{tabular}} & \multicolumn{1}{c|}{goatee}                                                    & sideburns                                                   \\ \hline
\multicolumn{1}{c|}{mustache}                                              & \multicolumn{1}{c|}{\begin{tabular}[c]{@{}l@{}}heavy\\ makeup\end{tabular}}    & \multicolumn{1}{c|}{\begin{tabular}[c]{@{}l@{}}wearing\\ earrings\end{tabular}} & \multicolumn{1}{c|}{\begin{tabular}[c]{@{}l@{}}wearing\_hat\end{tabular}}       & \multicolumn{1}{c|}{\begin{tabular}[c]{@{}l@{}}wearing\\ lipstick\end{tabular}} & \multicolumn{1}{c|}{\begin{tabular}[c]{@{}l@{}}wearing\\ necklace\end{tabular}}   & \multicolumn{1}{c|}{\begin{tabular}[c]{@{}l@{}}wearing\\ necktie\end{tabular}} & \begin{tabular}[c]{@{}l@{}}wearing\\ mask\end{tabular}      \\ \hline
\multicolumn{8}{c}{\textbf{(b) Action Attributes}} \\ \hline
\multicolumn{1}{c|}{blow}                                                  & \multicolumn{1}{c|}{chew}                                                      & \multicolumn{1}{c|}{close\_eyes}                                                & \multicolumn{1}{c|}{cough}                                                       & \multicolumn{1}{c|}{cry}                                                        & \multicolumn{1}{c|}{drink}                                                        & \multicolumn{1}{c|}{eat}                                                       & frown                                                       \\ \hline
\multicolumn{1}{c|}{gaze}                                                  & \multicolumn{1}{c|}{glare}                                                     & \multicolumn{1}{c|}{head\_wagging}                                              & \multicolumn{1}{c|}{kiss}                                                        & \multicolumn{1}{c|}{laugh}                                                      & \multicolumn{1}{c|}{listen\_to\_music}                                            & \multicolumn{1}{c|}{look\_around}                                              & make\_a\_face                                               \\ \hline
\multicolumn{1}{c|}{nod}                                                   & \multicolumn{1}{c|}{play\_instrument}                                          & \multicolumn{1}{c|}{read}                                                       & \multicolumn{1}{c|}{shake\_head}                                                 & \multicolumn{1}{c|}{shout}                                                      & \multicolumn{1}{c|}{sign}                                                         & \multicolumn{1}{c|}{sing}                                                      & sleep                                                       \\ \hline
\multicolumn{1}{c|}{smile}                                                 & \multicolumn{1}{c|}{smoke}                                                     & \multicolumn{1}{c|}{sneeze}                                                     & \multicolumn{1}{c|}{sneer}                                                       & \multicolumn{1}{c|}{sniff}                                                      & \multicolumn{1}{c|}{talk}                                                         & \multicolumn{1}{c|}{turn}                                                      & weep                                                        \\ \hline
\multicolumn{1}{c|}{whisper}                                               & \multicolumn{1}{c|}{wink}                                                      & \multicolumn{1}{c|}{yawn}                                                       & \multicolumn{1}{c|}{}                                                            & \multicolumn{1}{c|}{}                                                           & \multicolumn{1}{c|}{}                                                             & \multicolumn{1}{c|}{}                                                          &                                                             \\ \hline
\multicolumn{8}{c}{\textbf{(c) Emotion Attributes}} \\ \hline
\multicolumn{1}{c|}{neutral}                                               & \multicolumn{1}{c|}{anger}                                                     & \multicolumn{1}{c|}{contempt}                                                   & \multicolumn{1}{c|}{disgust}                                                     & \multicolumn{1}{c|}{fear}                                                       & \multicolumn{1}{c|}{happy}                                                        & \multicolumn{1}{c|}{sadness}                                                   & surprise                                                    \\ \hline
\end{tabular}}
\label{tbl:attr_names}
\end{table*}

The complete list of all the attributes is reported in Table \ref{tbl:attr_names}. 

\end{document}